\documentclass[12pt]{article}
\usepackage{amsmath}
\usepackage{amsfonts}
\usepackage{algorithmic}
\usepackage{algorithm}
\usepackage{array}
\usepackage[caption=false,font=normalsize,labelfont=sf,textfont=sf]{subfig}
\usepackage{textcomp}
\usepackage{stfloats}
\usepackage{verbatim}
\usepackage{etoolbox}

\usepackage{makecell}
\usepackage{multicol}
\usepackage{wrapfig}
\usepackage{times}
\usepackage{siunitx}
\usepackage{gensymb}
\usepackage{booktabs}
\usepackage{multirow}
\usepackage{xspace}

\usepackage{xcolor}
\usepackage{todonotes}
\usepackage{adjustbox}
\usepackage{pdflscape}

\usepackage{bbm}
\usepackage{tikz}
\usepackage{enumitem}
\setlist[itemize]{leftmargin=2em}
\setlist[enumerate]{leftmargin=2em}
\usepackage{hyperref}
\hypersetup{
    colorlinks=true,
    linkcolor=black,
    urlcolor=black,
    citecolor=black
}

\usepackage[CaptionAfterwards]{fltpage}
\usepackage{caption}

\usepackage{newtxtext,newtxmath}

\usepackage{graphicx}

\usepackage[letterpaper,margin=1in]{geometry}

\renewenvironment{abstract}
	{\quotation}
	{\endquotation}

\date{}

\makeatletter
\renewcommand{\fnum@figure}{\textbf{Figure \thefigure}}
\renewcommand{\fnum@table}{\textbf{Table \thetable}}
\makeatother

\usepackage{scicite}

\usepackage{url}

\newcommand{\mysection}[1]{\vspace{-0mm}\section*{#1}\vspace{-0mm}}
\newcommand{\mysubsection}[1]{\vspace{-0mm}\subsection*{#1}\vspace{-0mm}}
\newcommand{\mysubsubsection}[1]{\vspace{-0mm}\subsubsection*{#1}\vspace{-0mm}}
\newcommand{\mysubsubsubsection}[1]{\vspace{2mm}\noindent\textbf{#1}\vspace{-0mm}}

\newcommand{\ie}{i.e., }

\def\scititle{
	Time as a Control Dimension in Robot Learning
}
\title{\bfseries \boldmath \scititle}

\author{
	Yinsen Jia$^{1}$,
	Boyuan Chen$^{1, 2, 3\ast}$ \and
	\small$^{1}$Department of Electrical and Computer Engineering, Duke University \and
        \small$^{2}$Department of Mechanical Engineering and Materials Science, Duke University \and
        \small$^{3}$Department of Computer Science, Duke University \and
	\small$^\ast$To whom correspondence should be addressed; Email: boyuan.chen@duke.edu.
}

\begin{document} 

\maketitle

\begin{abstract} \bfseries \boldmath

Temporal awareness plays a central role in intelligent behavior by shaping how actions are paced, coordinated, and adapted to changing goals and environments. In contrast, most robot learning algorithms treat time only as a fixed episode horizon or scheduling constraint. Here we introduce time-aware policy learning, a reinforcement learning framework that treats time as a control dimension of robot behavior. The approach augments policies with two temporal signals, the remaining time and a time ratio that modulates the policy's internal progression of time, allowing a single policy to regulate its execution strategy across temporal regimes. Across diverse manipulation tasks including long-horizon manipulation, granular-media pouring, articulated-object interaction, and multi-agent coordination, the resulting policies adapt their behavior continuously from dynamic execution under tight schedules to stable and deliberate interaction when more time is available. This temporal awareness improves efficiency, robustness under sim-to-real mismatch and disturbances, and controllability through human input without retraining. Treating time as a controllable variable provides a new framework for adaptive and human-aligned robot autonomy.

\end{abstract}


\mysection{Introduction}
\noindent
Time is a fundamental dimension of intelligent behavior that governs how actions are paced, sequenced, and coordinated~\cite{baer1908auffassung, poppel1978time, zimbardo2014putting, maniadakis2014time}. Humans naturally adjust their tempo depending on context. When urgency increases, movements accelerate and exploit momentum; when precision is required, actions slow down to stabilize contact and reduce uncertainty. Timing also enables coordination between agents by aligning actions with shared schedules. For robots interacting with the physical world, comparable temporal competence directly affects efficiency, adaptability, safety, predictability, and collaboration.

Despite major advances in robot learning, most learned control policies remain effectively blind to time~\cite{hu2025flare, yang2024anyrotate, huang2021generalization}. Standard reinforcement learning (RL) typically optimizes cumulative reward under a fixed training horizon, without explicit mechanisms for allocating behavior under varying temporal requirements. As a result, policies that achieve high task success may still exhibit undesirable temporal structures. Robots often execute redundant motions, remain overly cautious when urgency is high, or adopt ``idle--then--rush'' strategies that delay progress early and compensate with abrupt dynamics near task completion~\cite{petrenko2023dexpbt, singla2024sapg, narang2022factory, tang2023industreal}. 
Such behaviors become particularly problematic in dynamic environments~\cite{liu2024robot, beigomi2024towards}, collaborative scenarios~\cite{umbrico2023human, maniadakis2017time, maniadakis2020time, li2024act, jia2024dynamic, liu2022unified}, and safety-critical deployments~\cite{rahman2021machine, gu2023human}, where timing itself forms part of the task specification.

A natural intuition is that cautious behavior can be obtained by simply slowing down a policy, for example by retiming trajectories or scaling control commands. 
However, many tasks in robotics such as manipulation involve complex contact dynamics in which instability arises from interaction modes rather than speed alone. Naively slowing down a trajectory may preserve the same unstable contact sequence while reducing responsiveness to disturbances. Moreover, simple temporal scaling provides no principled way to satisfy a desired completion time or adapt when schedules change. In practice, effective temporal reasoning requires policies to change strategy as temporal budgets vary. Under tight schedules the robot may exploit dynamic behaviors such as sliding or throwing to minimize duration, whereas under relaxed schedules it can invest additional time in stabilizing contacts and improving reliability.

Several lines of prior work have incorporated temporal considerations into robotics and reinforcement learning. Temporal scheduling frameworks coordinate agents under predefined timing constraints in collaborative robotics and human--robot interaction~\cite{wilcox2013optimization, rea2019human, cini2021relevance}. Other approaches incorporate temporal signals to compensate for delays or manage execution order in distributed robotic systems~\cite{gutierrez2018time, kagawa2017study, mastrogiovanni2012semantic, grass2018improved, baek2022reinforcement, jin2017improving}. Within reinforcement learning theory, work on time limits in Markov decision processes analyzes how episode truncation affects value estimation and shows that remaining time may need to be included in the state to preserve the Markov property~\cite{pardo2018time}. Related formulations incorporate time into value functions or discounting schemes to maintain temporal consistency across varying step durations~\cite{kim2023time, nhu2025time}. These studies treat time primarily as a structural variable required for learning correctness. They do not aim to make time a controllable dimension that shapes how a single policy behaves across different temporal regimes.

In this work we explore a different perspective. We hypothesize that temporal competence can emerge if time is incorporated not only as an observation but also as a controllable signal that influences how behavior unfolds. Under this view, time becomes an internal decision variable that modulates trade-offs between speed, stability, and responsiveness. The resulting policy can adapt continuously across temporal regimes, completing tasks punctually under tight schedules while adopting more cautious strategies when additional time is available.

\begin{FPfigure}
    \centering
    \includegraphics[width=\textwidth]{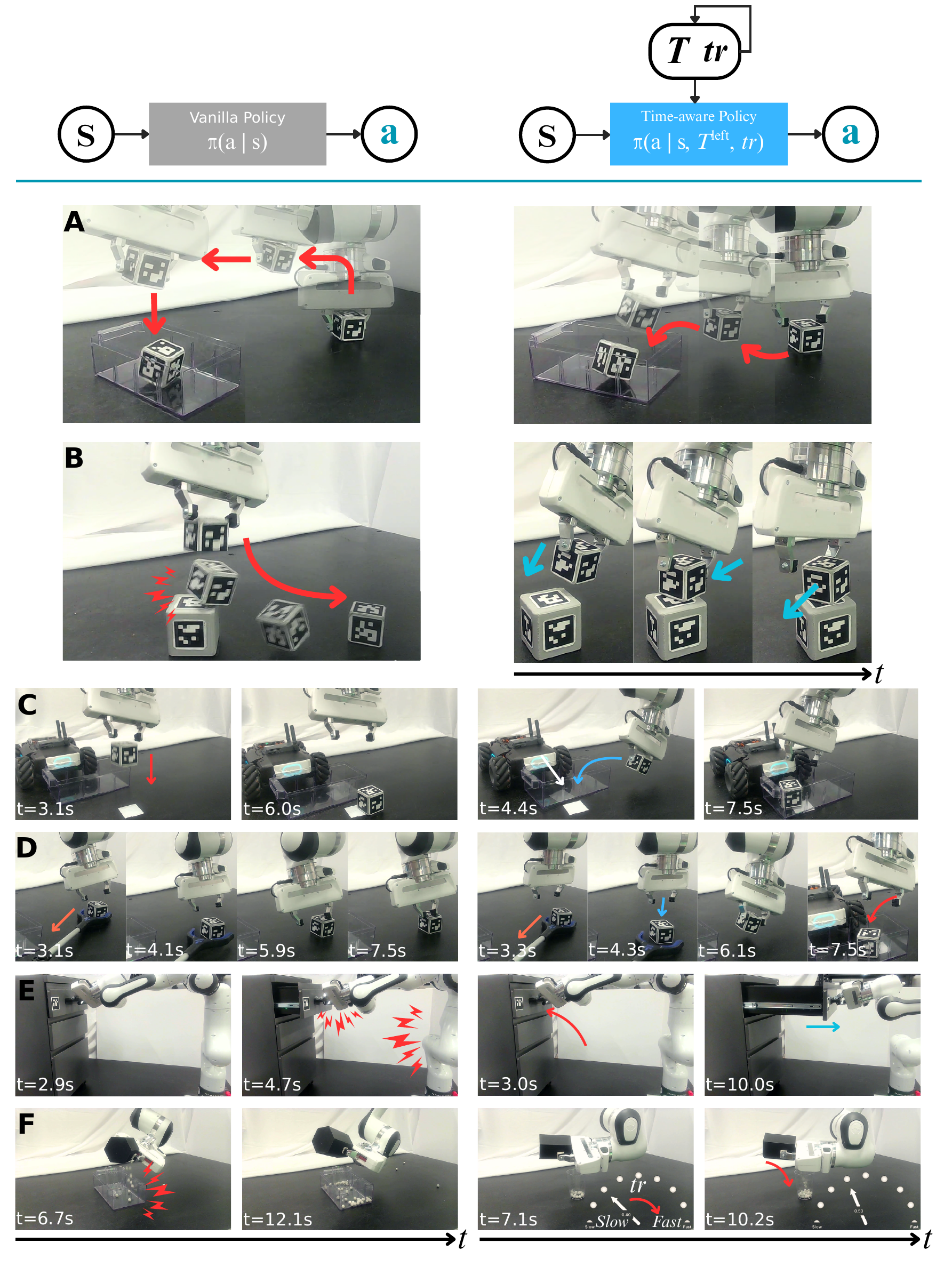}
    \caption{
        \textbf{Time-aware policy enables adaptive, punctual, robust, and controllable manipulation behaviors.}
        A standard policy $\pi(a \mid s)$ lacks temporal context. 
        Our time-aware policy $\pi(a \mid s, T^{\text{left}}, tr)$ conditions actions on remaining time $T^{\text{left}}$ and a time-ratio command $tr$ that modulates the internal perception of time.
        (A) Under tight time constraints the policy discovers dynamic strategies such as sliding or throwing that improve efficiency.
        (B) When additional time is available it adopts careful behaviors that stabilize contact and reduce sim-to-real degradation.
        (C) Predictable timing enables coordination in multi-agent collaboration.
        (D) Following perturbations the policy re-establishes contact and accelerates to meet the deadline.
        (E) Temporal modulation enables different execution phases within a single policy.
        (F) Real-time adjustment of $tr$ provides an intuitive interface for switching between aggressive and cautious modes.
    }
    \label{fig:teaser}
\end{FPfigure}

To realize this idea we introduce Time-Aware Policy Learning, a reinforcement learning framework that treats time as a control dimension in robot policy learning. We begin from the standard reinforcement learning objective
\begin{equation}
J_R(\pi) = \mathbb{E}\!\left[\sum_{t=0}^{\infty}\gamma^t r(s_t,a_t,s_{t+1})\right],
\end{equation}
where $s_t$ and $a_t$ denote the state and action at time step $t$, $r(\cdot)$ is the reward function, and $\gamma$ is the discount factor. While this formulation optimizes task success, it does not explicitly encourage efficient or temporally adaptive behavior.

Our approach augments the policy with two temporal signals. The resulting time-aware policy takes the form
\begin{equation}
\pi(a_t \mid s_t, T_t^{\mathrm{left}}, tr_t),
\end{equation}
where $T_t^{\mathrm{left}}$ denotes the remaining time available to complete the task and $tr_t$ is a time-ratio signal that modulates the policy's internal perception of time. Large values of $tr_t$ correspond to faster internal time progression and encourage time-efficient behaviors when urgency is high, whereas smaller values promote slower and more stable execution when additional time is available.

\mysubsubsection{Temporal calibration through time-optimal policy learning}

Learning temporally adaptive behavior requires calibrating the feasible time scale of a task. If temporal budgets are chosen arbitrarily, the policy may be required to finish faster than physically possible or may satisfy deadlines while wasting early time. We therefore first learn a time-optimal policy. Rather than training directly from scratch on a minimum-time objective, which is difficult for sparse-reward and long-horizon manipulation, we use a curriculum strategy. 
We first train a vanilla policy to solve the task reliably, and then continue training to emphasize temporal efficiency (Method:~\hyperref[sec:retrieve_timeoptimal]{Learning the temporal lower bound}). Executing the resulting time-optimal policy $\pi^{*}_{\mathrm{time}}$ over randomized task configurations yields an empirical temporal lower bound
\begin{equation}
T_{\min}(\phi) = 
\mathbb{E}\!\left[T^{\mathrm{succ}} \mid \pi^{*}_{\mathrm{time}}, \phi \right],
\end{equation}
where $T^{\mathrm{succ}}$ denotes task completion time and $\phi$ denotes the task configuration. This quantify establishes the reference for a feasible time window within which temporal modulation can occur.

\mysubsubsection{Learning time-aware policies.}

\begin{figure}
    \centering
    \includegraphics[width=\linewidth]{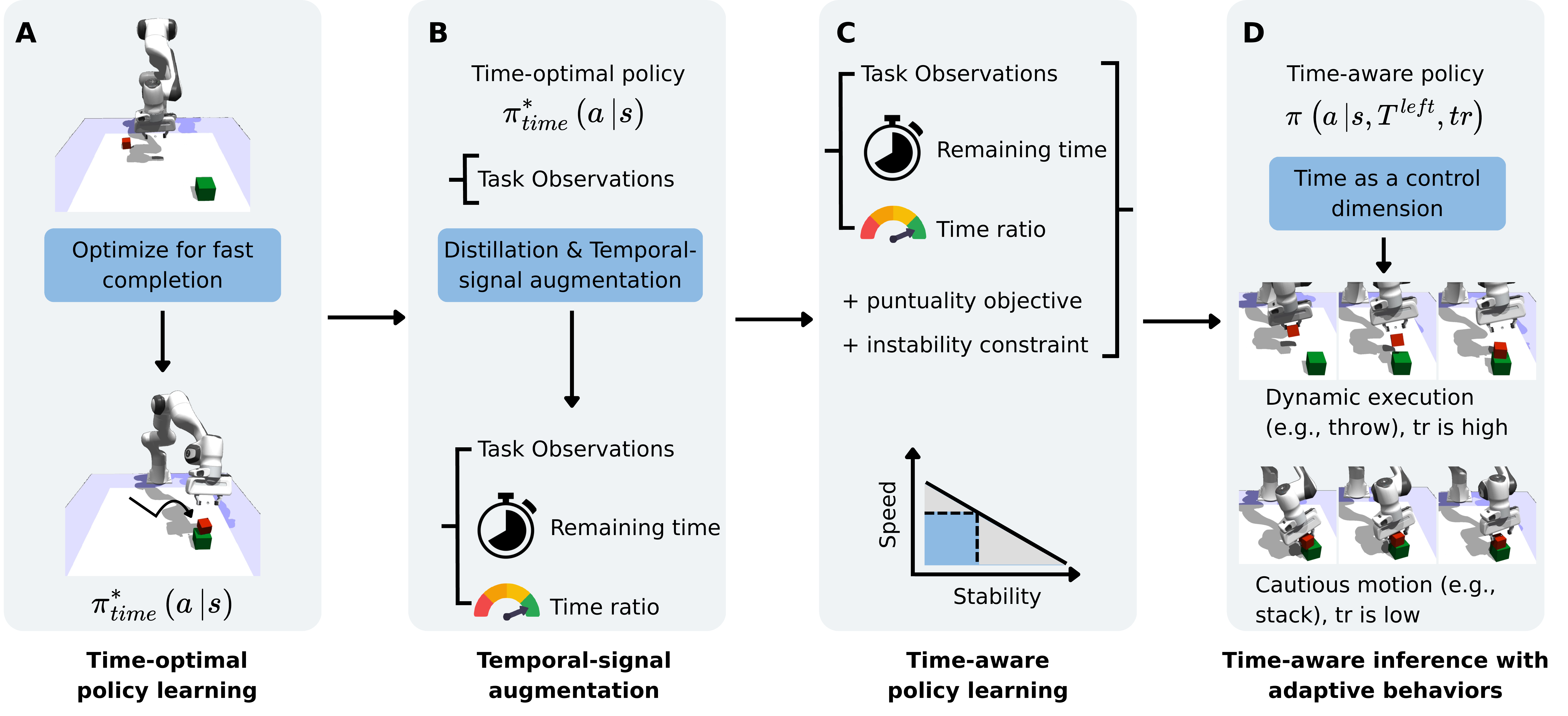}
    \caption{
        \textbf{Overview of time-aware policy learning.}
        (A) A time-optimal policy is first learned to estimate a temporal lower bound for the task.
        (B) Remaining time and time-ratio signals are appended to the policy observation space, and the time-optimal policy is distilled into this augmented policy.
        (C) The augmented policy is further trained with punctuality objectives and instability constraints, so that it adapts its behavior across schedules while using time effectively.
        (D) During deployment the time-ratio parameter modulates execution behavior without retraining.
    }
    \label{fig:method_overview}
\end{figure}

Time-aware policy learning proceeds in two stages. The aforementioned time-optimal policy provides both a feasible temporal reference and a fast baseline behavior (Fig.~\ref{fig:method_overview}A, Method:~\hyperref[sec:retrieve_timeoptimal]{Learning the temporal lower bound}). First, we augment the policy input with the temporal observations $(T_t^{\mathrm{left}}, tr_t)$. Directly appending these inputs and continuing RL training can destabilize learning and destroy the efficient behavior obtained in the first stage. To preserve the learned skill, we therefore perform behavior cloning from the time-optimal policy into an augmented student policy that receives temporal inputs but initially reproduces the teacher's behavior (Fig.~\ref{fig:method_overview}B). This stage teaches the policy to parse temporal signals without yet changing its strategy. Second, starting from this distilled policy, we continue RL training with explicit temporal objectives (Fig.~\ref{fig:method_overview}C, Method:~\hyperref[sec:retrieve_timeaware]{Time-aware policy optimization}) so that the same policy learns to adapt its behavior across different temporal regimes.

The time-aware policy maintains a remaining-time variable that evolves according to
\begin{equation}
T^{\mathrm{left}}_{t+1} = T^{\mathrm{left}}_t - \Delta t \cdot tr_t .
\label{eq:TimeAwareObj}
\end{equation}
Thus, the time ratio controls how quickly the policy's internal clock progresses. To encourage punctual completion, we penalize the terminal timing mismatch through the following formulations (Method:~\hyperref[sec:retrieve_timeaware]{Time-aware policy optimization}) to the total reward:
\begin{equation}
\delta_{\mathrm{time}} = \left|\frac{T^{\mathrm{left}}_K}{tr_K}\right|,
\end{equation}
where $K$ denotes the terminal step. This objective encourages the policy to finish near its scheduled time (i.e., punctuality) rather than merely complete the task eventually. The division by $tr_K$ is necessary because the remaining-time variable evolves according to the policy's internal clock. Dividing by $tr_K$ converts the residual remaining time back into the corresponding real-time mismatch, ensuring that the punctuality penalty is comparable across different temporal regimes.

Temporal conditioning alone, however, does not guarantee that extra time will be used effectively. A policy could still idle for much of an episode and act only near the end while remaining technically punctual. To prevent this behavior, we introduce an instability-based constraint during training. Scene instability is defined from the motion of manipulated objects, excluding the robot itself, and is constrained relative to the time ratio. This design makes instability an adaptive constraint based on the temporal context. When more time is available, the policy is encouraged to reduce instability and favor gentle, stable interactions. When time is limited, the policy is allowed to tolerate more dynamic motions to maintain punctuality. We incorporate this constraint through a constrained MDP formulation detailed in the Method (Method:~\hyperref[sec:cmdp_formulation]{Policy optimization}). Collectively, the punctuality objective and instability constraint cause the policy to use available time productively rather than simply waiting and rushing.

Across simulation and real-world manipulation tasks we show that explicit temporal conditioning enables robots to execute efficiently under tight schedules, behave cautiously when additional time is available, recover from perturbations while maintaining schedule consistency, and coordinate naturally with other agents. By treating time as a controllable dimension of policy learning, Time-Aware Policy Learning procvides a unified framework for adaptive, punctual, and human-steerable robot behavior.


\mysection{Results}

We evaluate Time-Aware Policy Learning in both simulation and real-world experiments to test five central capabilities: efficiency, punctuality, stability, adaptability, and controllability. Our goal is to determine whether treating time as a control dimension enables a single policy to change its execution strategy across different temporal requirements and environmental conditions, without retraining or reward redesign.

We study three representative manipulation domains spanning rigid-body interaction, granular media, and articulated objects with zero-shot sim-to-real transfer, and further evaluate the framework in multi-agent collaboration and human-in-the-loop control. Our experiments test whether explicit temporal control can improve how time is allocated, used, and reallocated throughout task executions.

\mysubsubsection{Experimental setup}
\label{sec:basic_setups}

We used a Franka Panda Research~3 arm for all manipulation tasks. As shown in Fig.~\ref{fig:model_architecture}, observations were first passed through a noise layer during training. The resulting noisy observations were then provided to the actor, which generated a 6-DoF relative end-effector pose and a binary gripper action at 20\,Hz. The relative pose command was converted into target joint positions using differential inverse kinematics. These positions were smoothed with an exponential moving average (EMA) filter and sent to a low-level joint impedance (JI) controller operating at 1\,kHz. For real-world evaluation, object poses were tracked using ArUco markers and an RGB camera. The minimum required time for each configuration were estimated by retrieving the five most similar simulated setups and averaging their successful completion time. Unless otherwise noted, each real-world condition was evaluated over at least ten trials and each simulation condition over 2,000 randomized task configurations.

\mysubsubsection{Tasks}

We selected tasks that capture several key challenges of real-world manipulation, including temporal precision, contact stability, dynamic adaptation, and coordination across agents. These task families are also commonly used to evaluate general-purpose robot control~\cite{petrenko2023dexpbt, makoviychuk2021isaac, zhu2020robosuite}.

\mysubsubsubsection{Cube placing and stacking}
This task instantiates long-horizon pick-and-place manipulation. In cube placing, the robot must pick up a cube from a random pose and place it into a box at a random location. In cube stacking, the robot must place one cube stably on top of another, which requires finer alignment and greater contact stability. A trial is considered successful when the cube is placed into the box or stably stacked on the target cube, and the end-effector is at least 5\,cm away from the source cube. Failure occurs if the contact force with the table exceeds 20\,N or the episode times out. We trained on cube stacking in simulation and evaluated both variants in the real world.

\mysubsubsubsection{Granular media pouring}
\label{sec:tasks_gms_pouring}
This task evaluates manipulation under high-contact and many-body dynamics. The robot must pour granular particles from a cup into a target container, which requires temporal modulation and continuous adaptation to flow behavior. In the real world, we used two variants: pouring into a large box and pouring into a small cup. Training in simulation used only a single granular particle, whereas testing scaled the number of grains up to 40 to assess generalization to unseen dynamic complexity. A trial succeeds if all particles are poured into the target container. Failure occurs if any particle is spilled, the cup is dropped, the table contact force exceeds 20\,N, or the episode times out. For the small-cup condition, an additional failure criterion prevents the target cup from toppling.

\mysubsubsubsection{Drawer opening}
This task evaluates articulated-object manipulation. The robot must grasp a drawer handle and pull it open to a displacement of 35\,cm. Success depends on both stable contact and compliant interaction, because overly rapid pulling can induce slip or destabilize the grasp. At the beginning of each episode, the cabinet pose is randomized in three translational degrees of freedom, and the arm is initialized at a random joint configuration. Failure occurs if any cabinet component experiences force greater than 20\,N or the episode times out.

\mysubsubsubsection{Multi-agent object delivery}
\label{sec:task_multiagent_box}
This collaborative task evaluates temporal coordination between agents. A robotic arm and a mobile vehicle must jointly deliver an object. The vehicle approaches a meeting location, pauses there, and then returns to a final location according to a pre-defined schedule. The arm must synchronize its placement motion with the vehicle’s arrival while maintaining stable object transfer. The task succeeds if the object is delivered within the scheduled temporal window.

\mysubsubsection{Metrics}

We evaluate performance using the following quantitative metrics:
\begin{itemize}
    \item \textbf{Success rate (\%):} fraction of trials completed successfully.
    \item \textbf{Elapsed time (s):} total elapsed time to complete an episode.
    \item \textbf{Time mismatch (s):} absolute difference between the scheduled and actual completion time in successful episodes.
    \item \textbf{Instability (m/s):} cumulative norm of manipulated-object velocity in successful episodes.
    \item \textbf{Maximum acoustic noise (amplitude ratio):} peak acoustic amplitude during manipulation, computed as $r_A(t)=\frac{A(t)}{A_{\max}}$, where $A_{\max}$ is the maximum recordable amplitude. We report $\max_t r_A(t)$ over the manipulation segment only. All audio measurements were collected in a controlled lab environment with no speech and minimal external activity.
\end{itemize}

\mysubsubsection{Baselines}

We compare our method against several baselines. For fairness, all policy-based baselines use the same infinite-horizon MDP formulation with timeout truncation.

\begin{itemize}
    \item \textbf{Vanilla policy}: trained with the original task dense reward and success reward, without temporal observations.
    \item \textbf{Time-optimal policy}: initialized from the vanilla policy and further trained with the time-optimal objective, without temporal observations.
    \item \textbf{Time-input policy}: trained with temporal observations, but without the punctuality objective or the instability constraint.
    \item \textbf{Time-dependent policy}: trained with temporal observations and the punctuality objective, but without the instability constraint.
    \item \textbf{Direct joints interpolation}: a non-learning baseline that slows the vanilla policy through joint-space interpolation. This tests whether simple temporal scaling can produce the adaptability of our method.
    \item \textbf{Constant time-ratio time-aware policy}: a variant of our method in which the time ratio is fixed from the beginning of the episode and not adjusted during inference.
    \item \textbf{Time-aware policy}: our full model trained with the complete time-aware learning procedure. The time ratio could be adjusted online during inference.
\end{itemize}

\begin{FPfigure}
    \centering
    \includegraphics[width=\linewidth]{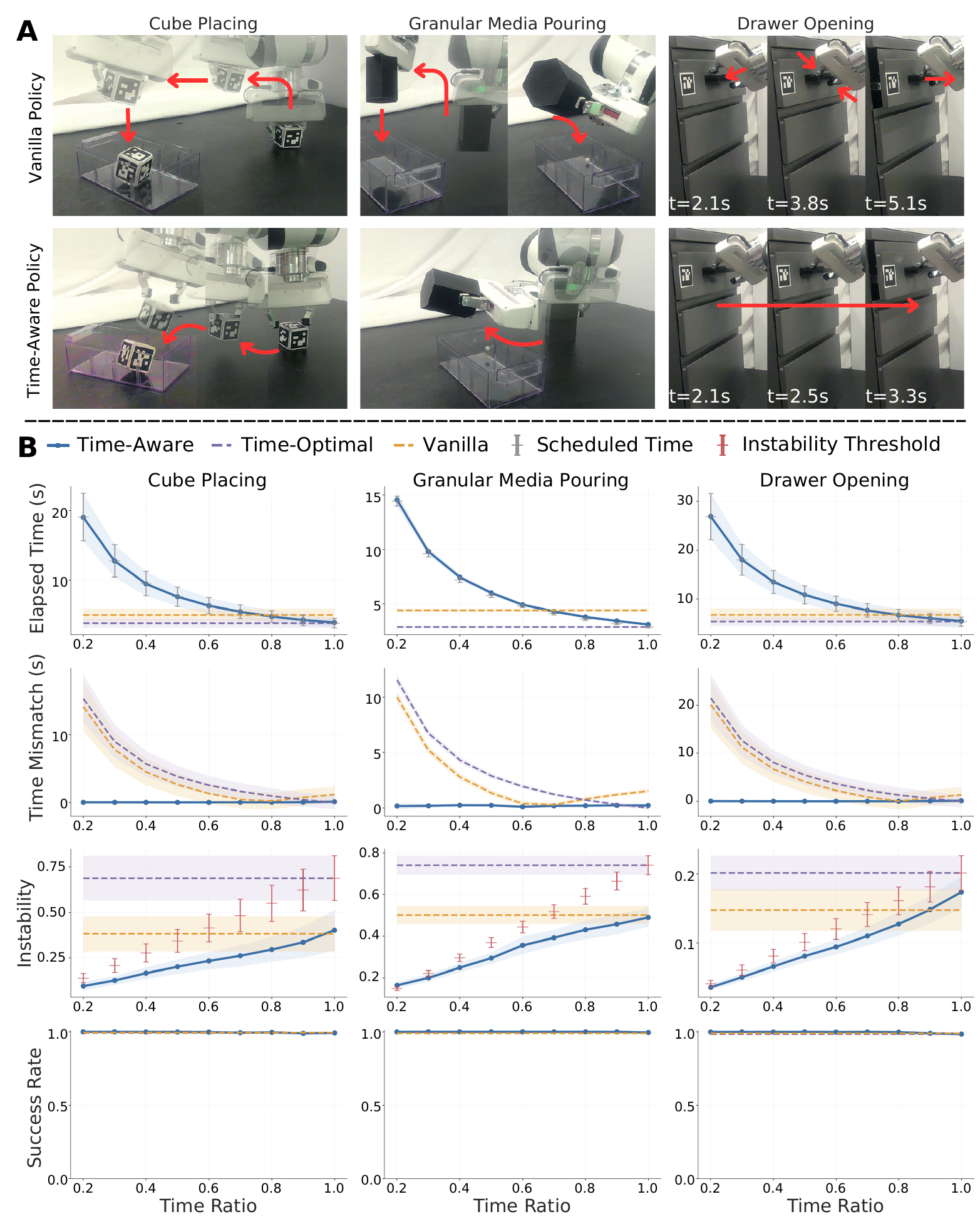}
    \caption{
        \textbf{Time awareness improves efficiency and punctuality.}
        \textbf{(A)} Real-world execution on three tasks. The vanilla policy follows largely sequential phases (approach, grasp, lift, place), whereas the time-aware policy under high time ratio ($tr=1$) exploits momentum and overlaps phases (e.g., slide-while-grasp, swing-to-pour, pre-tensioned drawer pull).
        \textbf{(B)} Simulation benchmarks across time-ratio settings.
        Elapsed time: completion time decreases as the time ratio increases. The time-aware policy is faster than the vanilla policy while adhering to the scheduled duration, and approaches the time-optimal policy at $tr=1$.
        Time mismatch: the time-aware policy maintains small completion-time mismatch across schedules.
        Instability: instability increases mainly at high time ratios enabled by our adaptive instability constraint, remains below the threshold, and stays substantially lower than the time-optimal policy at comparable efficiency.
        Success rate: the time-aware policy maintains near-100\% success across tasks and schedules.
        Explicit time awareness therefore enables efficient behavior under urgency and stable behavior when additional time is available.
    }
    \label{fig:Exp1_efficiency}
\end{FPfigure}

\mysubsubsection{Time awareness improves efficiency and punctuality}

We first evaluate whether treating time as a control dimension improves both execution efficiency and temporal precision. Under the minimum scheduled time ($tr=1.0$), the time-aware policy completed all three real-world manipulation tasks 18--48\% faster than the vanilla baseline (Table~\ref{tab:minimum_scheduled_time}). These efficient behaviors originate from the time-optimal policy used to estimate the temporal lower bound during training. The time-aware policy inherits these strategies through distillation and preserves them when the time ratio is set to $tr=1.0$. This ensures that the fastest feasible behavior is retained while enabling the same policy to adjust its execution when additional time is available. Movie S3 shows more qualitative demonstrations.

Qualitative behavioral differences reveal how explicit temporal reasoning reshapes manipulation strategies rather than simply accelerating motion. When operating under tight schedules, the policy reorganizes task phases to exploit momentum and overlap interaction stages. In cube placing, the vanilla policy largely followed a sequential pipeline of approach, grasp, lift, align, and release. The time-aware policy instead overlapped these phases, sliding while grasping and lifting while transporting, thereby exploiting inertia to produce throw-like placements into the target container (Fig.~\ref{fig:Exp1_efficiency}(A)). In granular pouring, the policy initiated particle flow during transport through a swing motion that shortened the effective pouring time. In drawer opening, it began pulling while closing the gripper, effectively pre-loading the interaction and eliminating idle delay after contact. Although these behaviors were learned in simulation, they transferred zero-shot to the real robot and consistently reduced completion time without sacrificing success rate. These observations indicate that time awareness enables policies to reorganize manipulation strategies under temporal pressure rather than merely executing the same strategy faster.

\begin{table}
	\centering
	\caption{\textbf{Vanilla policy versus time-aware policy under the minimum scheduled time.}
	We compare completion time for ten successful real-world trials under varying environment configurations.}
	\label{tab:minimum_scheduled_time}
	\small
	\begin{tabular}{lcc}
		\\
		\hline
		Task Name & Vanilla Policy & Time-aware Policy\\
		 & Elapsed Time (s) $\downarrow$ & Elapsed Time (s) $\downarrow$\\
		\hline
		Cube Placing & $6.24\pm0.53$ & $\mathbf{3.25\pm0.23}$\\
		Granular Media Pouring (Box) & $6.34\pm0.71$ & $\mathbf{4.21\pm0.57}$\\
		Drawer Opening & $8.53\pm0.14$ & $\mathbf{7.03\pm0.48}$\\
		\hline
	\end{tabular}
\end{table}

Large-scale simulation over 2,000 randomized configurations per task and time ratio confirmed the same pattern (Fig.~\ref{fig:Exp1_efficiency}(B)). The vanilla policy exhibited essentially fixed behavior and could not adapt to different schedules. In contrast, the time-aware policy tracked the target timing with an average mismatch below 0.6\,s across all time ratios. At $tr=1$, it used 25\% less time than the vanilla policy while maintaining the original success rate ($>99\%$). Relative to the time-optimal policy, it achieved 32\% lower instability at similar efficiency. As the time ratio decreased, the policy used the additional time to act more cautiously and kept instability below the designated threshold. These results show that explicit time awareness improves not only how quickly tasks are completed under urgency, but also how precisely and effectively time is used during execution.

\mysubsubsection{Adaptive stability and environmental robustness}
\label{sec:stability_tasks}

We next test whether explicit time awareness improves stability and generalization under environment changes not seen during training. To create realistic train vs. test mismatch, we modified each real-world setup to require more delicate and robust behavior. In cube placing, we replaced the target box with another cube to create cube stacking. In granular pouring, we replaced the box with a same-sized cup and increased the number of particles from 1 during training to 40 at test time. In drawer opening, we added 2\,kg to the drawer to simulate heavier and unseen loads. These changes altered contact dynamics, friction, and actuator load, producing substantial sim-to-real discrepancies.

Under these modified conditions, the vanilla policy degraded sharply across all tasks. In cube stacking, it repeatedly failed to align the cubes because of unmodeled friction and material properties, revealing a clear dynamics mismatch. In granular pouring, it retained the aggressive motions learned during training and spilled a substantial fraction of the particles. In drawer opening, it frequently pulled too rapidly; under the added drawer weight, the lag between commanded and realized motion accumulated and destabilized the impedance controller (Fig.~\ref{fig:Exp2_stability}(A)).

Adapting to such shifts conventionally requires laborious real-to-sim recalibration followed by policy fine-tuning. This process is expensive and brittle, and each additional change in dynamics may require another cycle of system identification and retraining. In contrast, we found that the time-aware policy adapted to all of these modified conditions simply by increasing the scheduled time.

\begin{table}
	\centering
	\caption{\textbf{Vanilla policy v.s. time-aware policy at 2$\times$ minimum scheduled time).} We compare performance on 10 real-world trials under modified test conditions using $2\times T^{\text{min}}$ (equivalently, $tr=0.5$).}
	\label{tab:double_scheduled_time}
	\begin{tabular}{lcccc}
		\\
		\hline
		Task Name & \multicolumn{2}{c}{Vanilla Policy} & \multicolumn{2}{c}{Time-aware Policy $(tr=0.5)$}\\
		 & Success Trials $\uparrow$ & Max Noise $\frac{A}{A_{max}}$ $\downarrow$ & Success Trials $\uparrow$ & Max Noise $\frac{A}{A_{max}}$ $\downarrow$\\
		\hline
		Cube Stacking & 1/10 & $1.0\pm0.0$ & \textbf{9/10} & $\mathbf{0.09} \pm \mathbf{0.04}$\\
		Pouring (40) & 0/10 & $1.0\pm0.0$ & \textbf{8/10} & $\mathbf{0.73} \pm \mathbf{0.15}$\\
		Drawer Opening & 6/10 & $0.53\pm0.14$ & \textbf{10/10} & $\mathbf{0.03} \pm \mathbf{0.01}$\\
		\hline
	\end{tabular}
\end{table}

Doubling the scheduled time for the time-aware policy immediately mitigated the observed failures without retraining or reward adjustment. As shown in Fig.~\ref{fig:Exp2_stability}(A) and Table~\ref{tab:double_scheduled_time}, the time-aware policy achieved 9/10 success in small cube stacking, 8/10 in granular pouring with 40 particles, and 10/10 in heavy drawer opening. In cube stacking, the policy carefully aligned the top cube before release. In granular pouring, it executed a slower, more sequential pour and even exhibited gentle shaking to release residual particles. In drawer opening, the extra time allowed it to secure a firm grasp and increase pulling force gradually, which stabilized the impedance control loop.

An additional practical benefit was a large reduction in manipulation noise. Careless robotic behaviors such as dropping, striking, or rapid contact transitions often generate excessive acoustic noise, yet directly rewarding quiet manipulation can be difficult. The time-aware policy naturally reduced maximum acoustic noise by over 90\% relative to the vanilla baseline across all three real-world tasks (Table~\ref{tab:double_scheduled_time}) when given more time ($tr = 0.5$). Its motions became noticeably quieter, which is particularly relevant for household and human-centered applications. See Movie~S4 for additional examples.

To quantify adaptation more systematically, we repeated these evaluations in simulation over 2,000 randomized configurations per condition. For cube stacking, we increased the restitution coefficients of both cubes from 0.1 to 1.0 ($\times 10$ times) to approximate the higher bounciness observed in reality. The vanilla policy achieved only 15.3\% success, whereas the time-aware policy reached 99.2\% success when given twice the scheduled time. In this case, allocating more time did not yield further gains. In granular pouring, the vanilla policy remained overly aggressive and spilled increasingly many grains as the cup filled (Fig.~\ref{fig:Exp2_stability}(B)). The time-aware policy instead reorganized its behavior according to the remaining time and time ratio. Moderately reducing the time ratio yielded slower, lower-velocity pouring that ensured success, whereas making the behavior too conservative caused some grains to remain in the cup. Thus, intermediate temporal regimes were optimal. Overly slow behavior can therefore be as suboptimal as overly fast behavior, underscoring the importance of time awareness as a control dimension (Fig.~\ref{fig:Exp2_stability}E).

\begin{FPfigure}
    \centering
    \vspace{-5mm}
    \includegraphics[width=\linewidth]{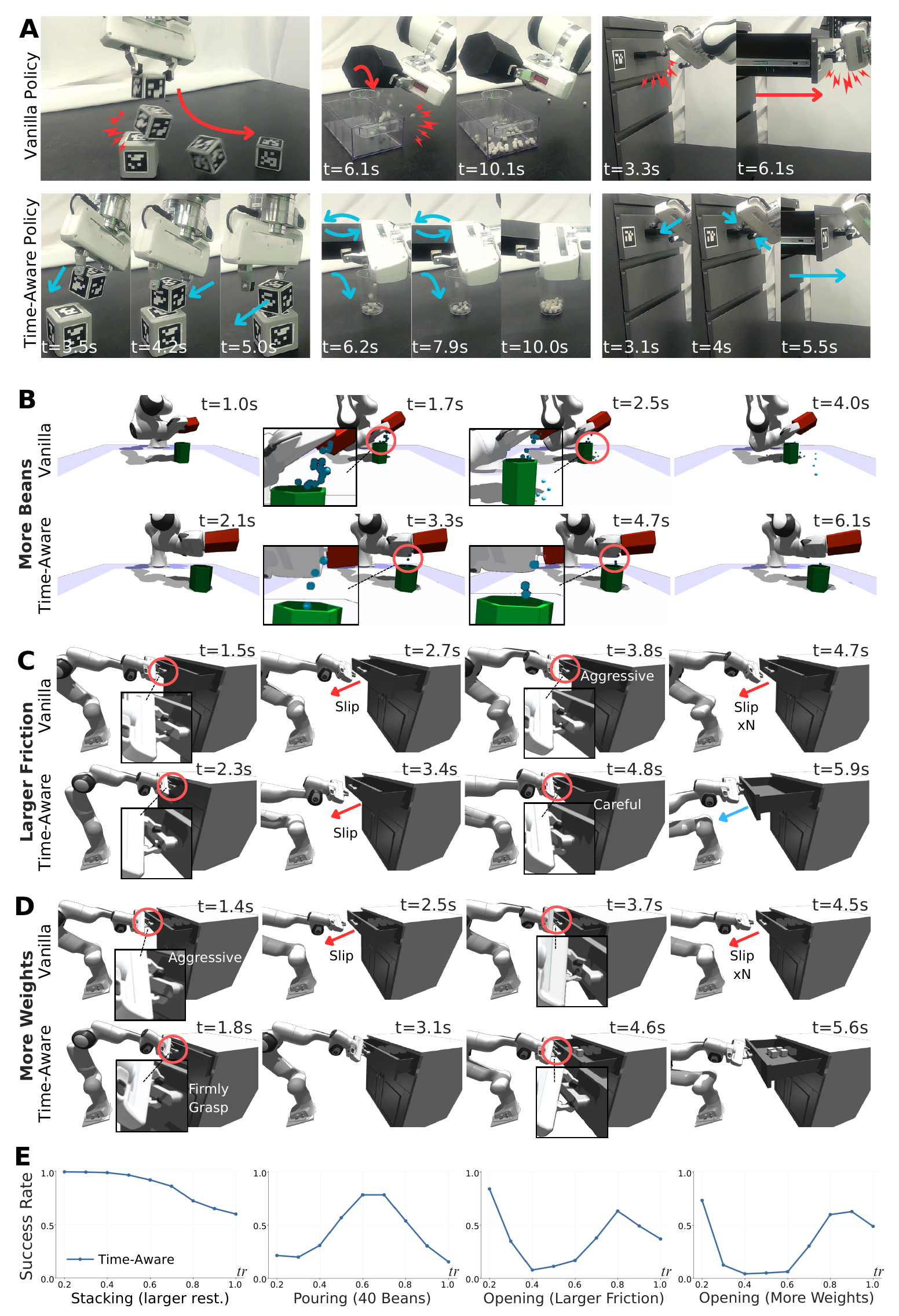}
    \caption{
        \textbf{Time awareness improves stability and robustness to unseen dynamics.}
        \textbf{(A)} With more time available, the time-aware policy adopts cautious behaviors that
        (i) mitigate sim-to-real gaps,
        (ii) generalize to unseen scenarios such as 40$\times$ more granular media than used during training,
        (iii) avoid aggressive motions that hit controller limits, and
        (iv) reduce manipulation noise.
        \textbf{(B)--(D)} Zero-shot generalization tests in simulation.
        For granular pouring, both policies were trained on one particle and evaluated on 40; the baseline remained aggressive and spilled substantially, whereas the time-aware policy slowed down appropriately.
        For drawer opening, both policies were trained on an empty drawer and tested under increased friction or added weight. The baseline slipped and failed repeatedly, whereas the time-aware policy adapted by grasping more firmly and pulling more carefully.
        \textbf{(E)} The time-aware policy outperforms the baseline across time ratios. No single fixed temporal regime performs best across tasks, indicating that neither purely aggressive nor purely conservative behavior is sufficient.
    }
    \label{fig:Exp2_stability}
\end{FPfigure}

For drawer opening, we examined two additional perturbations that commonly arise in deployment. First, increasing cabinet friction induced repeated slip events in the vanilla policy due to aggressive pulling (Fig.~\ref{fig:Exp2_stability}(C), Movie~S5). The time-aware policy instead used the extra time to maintain stronger grasp force and slower extension, and it could recover from minor slips. Interestingly, intermediate time ratios between $0.3$ and $0.7$ yielded the lowest performance in this task (Fig.~\ref{fig:Exp2_stability}(E)), unlike granular pouring, indicating that optimal temporal strategies depend on the manipulation domain. Second, increasing drawer weight again caused the time-aware policy to outperform the baseline by executing careful, progressive pulls and recovering from transient slip events (Fig.~\ref{fig:Exp2_stability}(D), Movie~S5). In both cases, the time ratio acted as an implicit regulator of contact stability.

\begin{figure}
    \centering
    \includegraphics[width=\linewidth]{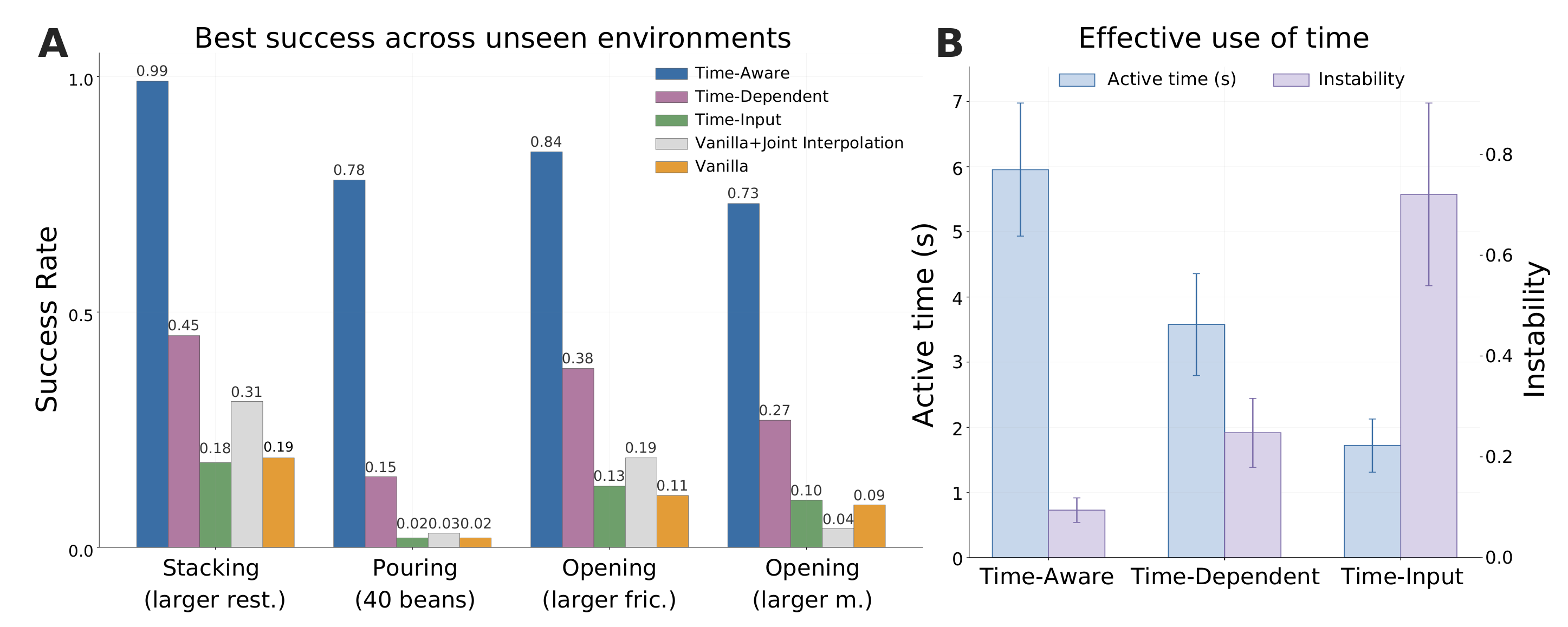}
    \caption{
        \textbf{Time awareness improves stability and robustness to unseen dynamics (continued).}
        \textbf{(A)} All four baselines fail to adapt to unseen environments that require cautious behavior, whereas the full time-aware policy succeeds across all modified tasks.
        \textbf{(B)} The time-dependent policy increases active manipulation time and reduces instability to some extent, but the full combination of punctuality and stability objectives improves both time utilization and overall stability further.
    }
    \label{fig:Exp2_stability_continue}
\end{figure}

We also tested whether simple temporal scaling could reproduce these gains. In the direct joints interpolation baseline, the vanilla policy's target joint command at each control step was split into three linearly interpolated intermediate configurations executed at the same 20\,Hz control frequency as our method. Although this slowed the motion, the resulting behavior remained qualitatively similar to the vanilla policy and achieved similar success rates (Fig.~\ref{fig:Exp2_stability_continue}(A)). The robot still displaced cubes during stacking, spilled particles during pouring, and slipped while opening the drawer. Temporal scaling alone therefore did not produce adaptive behavior.

To isolate the role of each training component, we compared the full model against the time-input and time-dependent baselines across all four task variants in simulation, reporting the best success rate over a grid search of time ratios (2,000 trials per task). The time-input policy, which observes temporal signals but optimizes neither punctuality nor stability, changed little across time ratios and performed similarly to the vanilla policy with joint interpolation. The time-dependent policy, which optimizes punctuality alone, improved performance relative to time-input but remained consistently below the full time-aware policy (Fig.~\ref{fig:Exp2_stability_continue}(A)). These results show that temporal observations alone are insufficient, and that the instability constraint is essential for converting extra time into meaningful improvements in behavior.

To quantify whether additional time was used productively, we defined ``active time'' as the duration during which the manipulated object, rather than the robot, was moving. Under identical scheduled times across all four task variants, the time-input policy exhibited short active time and high instability, indicating that it often ignored temporal observations and remained idle before acting. The time-dependent policy increased active manipulation and reduced instability to some extent, but still displayed idle--then--rush behavior. In contrast, the full time-aware policy increased active time for meaningful interaction by 62.2\%, reduced peak instability, and still met the scheduled completion time (Fig.~\ref{fig:Exp2_stability_continue}(B)). These ablations indicate that both punctuality and stability objectives are necessary to produce temporally adaptive and effective manipulation.

\mysubsubsection{Interpreting the role of temporal observations}
\label{sec:interp_temporal_obs}

\begin{figure}
    \centering
    \includegraphics[width=\linewidth]{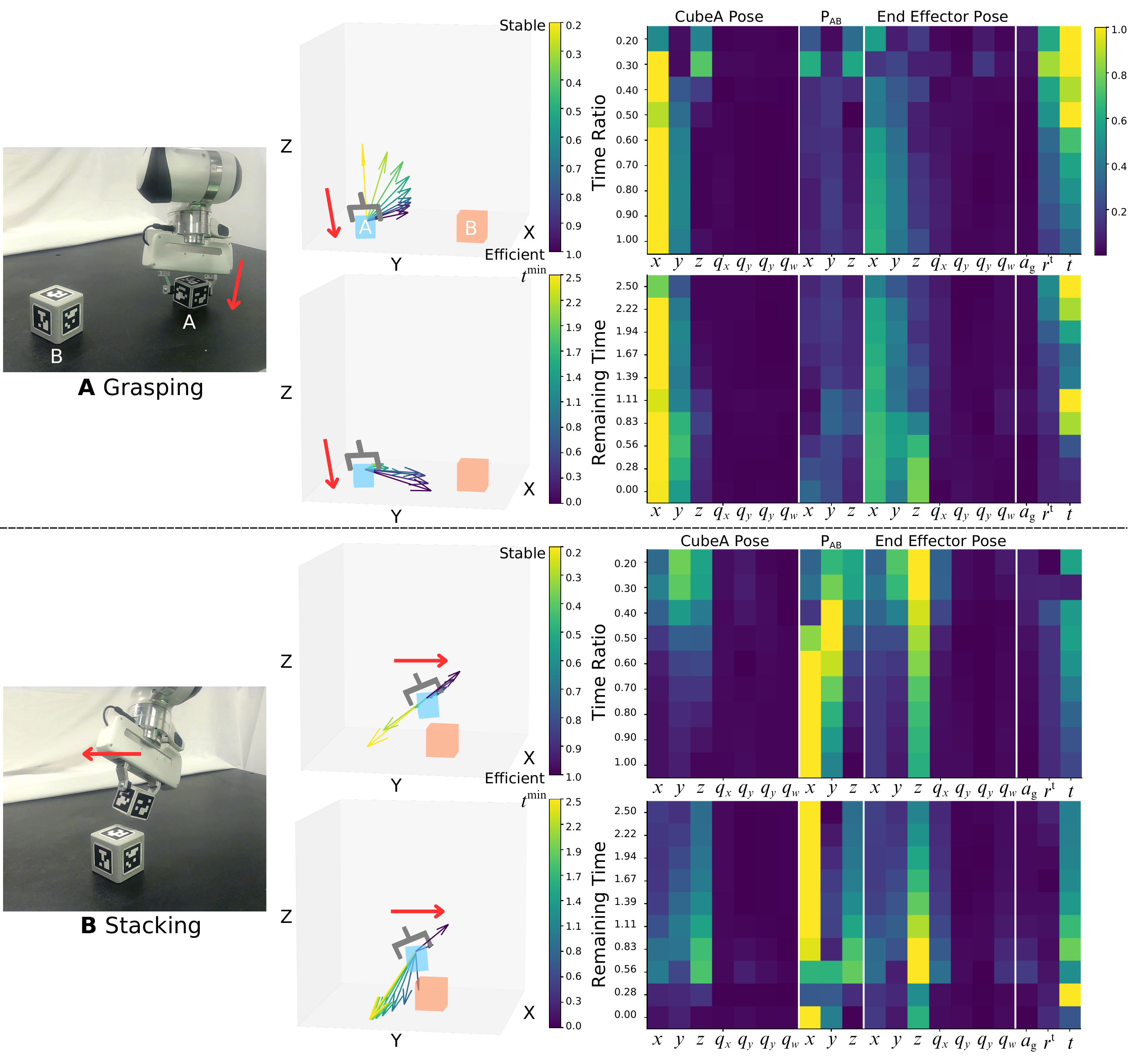}
    \caption{
        \textbf{Temporal observations modulate action generation and observation saliency.}
        We analyze the effect of temporal variables during two manipulation stages: \textbf{(A)} grasping and \textbf{(B)} stacking.
        Left: current real-world snapshots.
        Middle: actions under modified temporal observations, with other observations held fixed, visualized in simplified 3D space.
        Right: heatmaps showing changes in observation importance using normalized gradient magnitude, \(\left\| \frac{\partial \pi(a \mid s, T^{\text{left}}, tr)}{\partial s} \right\|\), scaled to $[0,1]$.
        Within each stage, the upper row varies the time ratio and the lower row varies the remaining time.
        Temporal observations influence both the selected action and which state variables dominate the policy’s decision.
    }
    
    \label{fig:time_analysis}
\end{figure}

To understand how temporal signals shape decision making, we examined the policy's responses to controlled changes in remaining time and time ratio. We hypothesized that temporal inputs do not merely modulate action magnitude, but also alter which observations the policy prioritizes. Figure~\ref{fig:time_analysis} shows two representative frames from cube stacking corresponding to grasping and stacking.

During grasping (Fig.~\ref{fig:time_analysis}(A)), smaller time ratios produced counter-directed actions that decelerated the motion for more stable grasping, whereas larger time ratios generated forward-directed sliding actions toward the target cube. Similarly, shorter remaining time led to more aggressive actions, reflected by arrows of larger magnitude, whereas longer remaining time produced gentler and more deliberate trajectories. Under tight schedules, the policy focused mainly on local variables such as \texttt{cubeA pos X, Y} and end-effector position. With more available time, it additionally attended to \texttt{cubeA pos Z}, translations between the cubes, time ratio, and remaining time, supporting more deliberate and stable behavior.

During stacking (Fig.~\ref{fig:time_analysis}(B)), increasing remaining time, or equivalently lowering the time ratio, slowed the motion and improved alignment before release. Decreasing remaining time produced fast, throw-like behaviors. Under urgent schedules, the policy concentrated on a narrow set of task-critical variables to finish quickly. Under relaxed schedules, it incorporated a broader set of spatial and temporal relations, including \texttt{cubeA} pose, inter-cube translation, end-effector pose, and remaining time. These analyses suggest that temporal awareness changes not only how the robot moves, but also how it allocates perceptual attention during action generation.

\begin{figure}
    \centering
    \includegraphics[width=\linewidth]{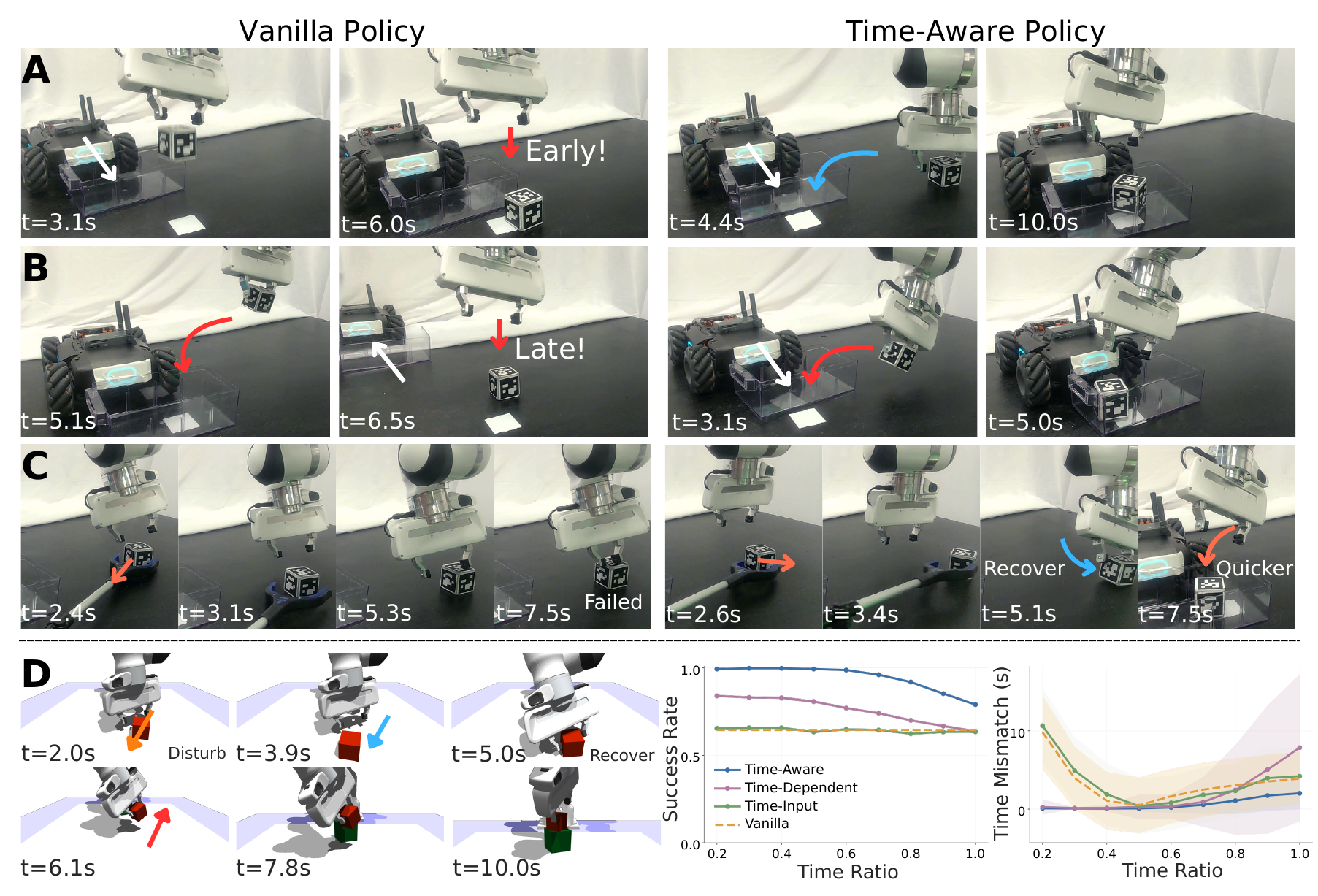}
    \caption{
        \textbf{Time-aware policy enables punctual, adaptive, and robust multi-agent collaboration.}
        A robot arm must deliver an object to a mobile vehicle by synchronizing with the vehicle's scheduled arrival at a meeting point.
        \textbf{(A)} With a 10\,s schedule, the vanilla policy releases too early, whereas the time-aware policy remains punctual.
        \textbf{(B)} With a 5\,s schedule, the vanilla policy again misses the handover time, while the time-aware policy adapts successfully.
        \textbf{(C)} Under disturbances, the time-aware policy recovers and preserves coordination, whereas the baseline fails.
        \textbf{(D)} Similar disturbances in cube stacking show that the time-aware policy recovers quickly and still completes near schedule.
    }
    \label{fig:Exp3_multi_agent}
\end{figure}

\mysubsubsection{Collaboration and resiliency in multi-agent settings}
\label{sec:Exp3_multi_agent}

We next ask whether explicit time awareness can support coordination and recovery in multi-agent tasks. In the object delivery task, a robotic arm and a mobile vehicle must complete a handover under a shared temporal schedule. This setting raises three challenges: each agent's completion time is uncertain, online adaptation is required when timing changes, and failed recovery by one agent can delay the other.

We varied the vehicle's travel duration between 5\,s and 10\,s per episode. The vanilla policy was highly sensitive to these schedule changes and often released the object either too early or too late, causing handover failure (Fig.~\ref{fig:Exp3_multi_agent}(A,B)). In contrast, the time-aware policy explicitly reasoned about the remaining time at each control step and adapted online when the arrival time changed during execution. Across these 5--10\,s schedules, the time-aware policy succeeded in all 10/10 trials with an average timing deviation of 0.43\,s, whereas the vanilla policy succeeded in only 2/10 trials (Table~\ref{tab:box_delivery_quant}).

We then tested recovery under perturbation. Fixing the delivery deadline at 7.5\,s, we applied random external pushes to the box while the gripper approached it (Fig.~\ref{fig:Exp3_multi_agent}(C)). The vanilla policy failed in all disturbed trials, reflecting a strong dependence on disturbance-free trajectories learned during training. By contrast, the time-aware policy re-established contact through a smooth re-grasp and then accelerated to compensate for the lost time, achieving 7/10 successful recoveries with an average timing deviation of 0.45\,s (Table~\ref{tab:box_delivery_quant}).

To analyze this mechanism in a more controlled setting, we reproduced a similar disturbance in cube stacking in simulation. When the gripper was within 5\,cm of the source cube, we applied a planar 10\,N force to the cube for one simulation step, and required the policy to complete the task within the original schedule. The vanilla policy was brittle: its success rate dropped by 35.1\%, and it often became trapped in failed re-grasp attempts. The time-input policy showed similar behavior, indicating that temporal observations alone were insufficient. The time-dependent policy improved recovery at small time ratios, but under high urgency ($tr>0.7$) required substantially more time to recover and accumulated large mismatch. The full time-aware policy achieved the lowest timing mismatch across all time ratios, and when $tr<0.5$ its final mismatch remained below 0.32\,s (Fig.~\ref{fig:Exp3_multi_agent}(D)).

\begin{table}
	\centering
    \caption{\textbf{Evaluation of the multi-agent box delivery task.}
    Success trials indicate the number of successful attempts out of 10. Time difference is measured over successful trials only.}
	\label{tab:box_delivery_quant}
	\small
	\begin{tabular}{lcccc}
		\\
		\hline
		Box Delivery & \multicolumn{2}{c}{Vanilla Policy} & \multicolumn{2}{c}{Time-aware Policy}\\
		 & Success Trials $\uparrow$ & Time Diff (s) $\downarrow$ & Success Trials $\uparrow$ & Time Diff (s) $\downarrow$\\
		\hline
		Varied Time (5--10s) & 2/10 & - & \textbf{10/10} & $\mathbf{0.43} \pm \mathbf{0.47}$\\
		Random Disturbance & 0/10 & - & \textbf{7/10} & $\mathbf{0.45} \pm \mathbf{0.89}$\\
		\hline
	\end{tabular}
\end{table}

These results reveal two complementary strategies induced by explicit time awareness: cautious recovery, in which the policy slows down to re-establish stable contact when time is available, and temporal compensation, in which it accelerates afterward to restore the original schedule. Time awareness therefore provides a unified mechanism for both coordination and resilience.

\mysubsubsection{Human-in-the-loop temporal control for real-time behavior adaptation}
\label{sec:online_interface_control}

A major practical advantage of time awareness is that the same policy can be modulated online to suit different user preferences or task contexts. In industrial settings, users may prioritize throughput and therefore prefer the shortest feasible schedule. In household or collaborative settings, quieter and safer operation may be more desirable. Because the time-aware policy is explicitly grounded in temporal control signals, its execution tempo can be adjusted online without retraining.

We study two complementary control modes. In heuristic stage-wise control, a manipulation task is divided into semantic stages such as approach, grasp, transport, and placement, and each stage is assigned a different time ratio. This allows the same policy to move aggressively when contact is not critical and cautiously when precision is essential. In online interface control, the user directly adjusts the time ratio through a simple interface, such as a keyboard or slider, to shape the robot's behavior in real time.

\begin{figure}
    \centering
    \includegraphics[width=\linewidth]{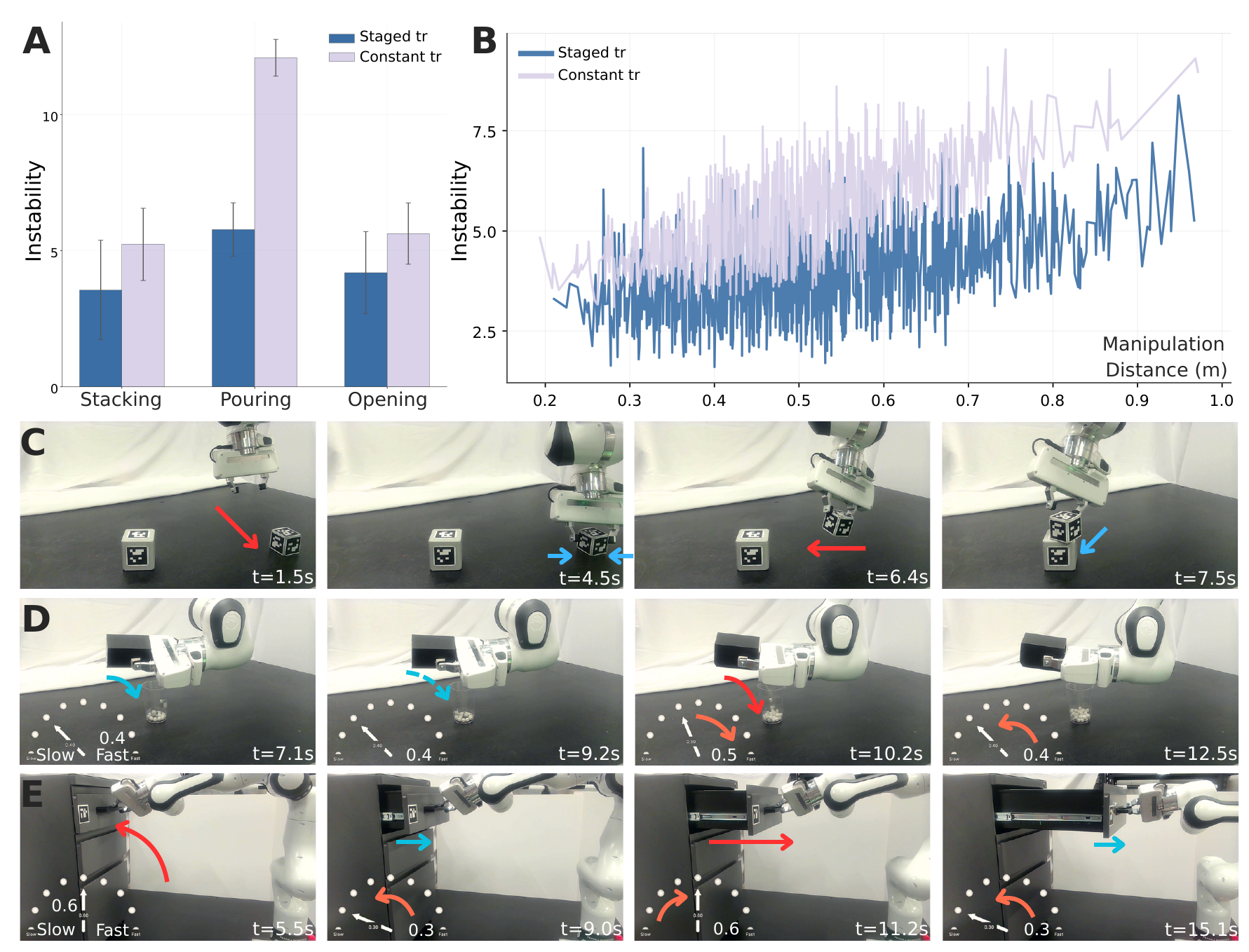}
    \caption{
        \textbf{Explicit time awareness enables stage-wise control and interactive online adjustment.}
        \textbf{(A)} Stage-wise control allocates more time to contact-rich stages and yields markedly lower instability across tasks.
        \textbf{(B)} The benefit of stage-wise control increases with task horizon. For cube stacking, we defined manipulation distance as the initial gripper-to-cube distance plus cube-to-target distance. As this distance increased, stage-wise control achieved larger instability reductions under identical scheduled time.
        \textbf{(C)} Example cube-stacking episode using fast transport and slow contact-rich execution.
        \textbf{(D, E)} Online adjustment of the time ratio corrects behavior under unmodeled conditions in granular pouring and drawer opening. In granular pouring, increasing the time ratio overcame media sticking from overly gentle actions. In drawer opening, reducing the time ratio during initial pulling, then gradually increasing it, mitigated initial lag and stabilized the low-level controller. This interface enables robot adaptation to new scenarios through high-level human feedback.
    }
    \label{fig:Exp3_staged_online}
\end{figure}

For heuristic stage-wise control, we assigned each stage a binary mode, efficient (E) or stable (S), together with a fraction of the overall temporal budget. The associated time ratios were then computed so that the total schedule remained unchanged while stable phases received smaller time ratios and efficient phases larger ones (Method:~\hyperref[sec:stage_wise_control]{Heuristic stage-wise control details}; Table~\ref{tab:Staged_wise_control}).

Real-world demonstrations for all three manipulation tasks are shown in Movie~S7. Figure~\ref{fig:Exp3_staged_online}(C) shows a representative cube-stacking episode in which a large time ratio during approach and transport produced faster motion, whereas a smaller ratio during grasping and stacking produced slower and more stable behavior. Across trials, the policy completed the task within the target schedule while lowering instability in the contact-rich phases.

Because instability is difficult to measure reliably in real-world scenes, we quantified stage-wise control in simulation. Under the same overall schedule, stage-wise control produced substantially lower instability than using a constant time ratio throughout the entire episode (Fig.~\ref{fig:Exp3_staged_online}(A)). This indicates that time is used more effectively when caution is allocated only to the phases that need it. Moreover, the advantage of stage-wise control grew with manipulation distance (Fig.~\ref{fig:Exp3_staged_online}(B)), because longer tasks provide more absolute time that can be reallocated to stability-critical phases.

We also tested direct user control of the time ratio. In granular pouring, an initially cautious policy caused particles to adhere to the cup; increasing the time ratio generated stronger motion and enabled successful pouring (Fig.~\ref{fig:Exp3_staged_online}(D)). In drawer opening, an initially aggressive pull saturated the joint velocity limit and destabilized control; reducing the time ratio early and increasing it later enabled smooth initiation followed by efficient completion (Fig.~\ref{fig:Exp3_staged_online}(E)). These examples show that time awareness provides a practical, human-interpretable interface for modulating learned robot behavior online.

Our results highlight the versatility of explicit temporal control. A single policy can accommodate pre-programmed stage priorities as well as real-time human feedback, enabling deployment in settings where user intent, environmental uncertainty, and safety requirements demand adaptive and human-aligned control.

\mysection{Discussion}

We introduced Time-Aware Policy Learning, a reinforcement learning framework that treats time as an explicit control dimension in robot policy learning. Rather than treating time as a fixed episode horizon or a scheduling constraint, the proposed framework embeds temporal signals directly into the policy representation and training objectives. Through the joint use of remaining time and a controllable time-ratio signal, the policy learns to regulate its execution strategy according to temporal context. The resulting policy preserves the efficient manipulation strategies inherited from the time-optimal policy while enabling continuous modulation of behavior across different temporal regimes.

Across both simulation and real-world experiments, this formulation produced a single policy capable of adapting its interaction strategy under varying time budgets without retraining or reward redesign. When time is scarce, the policy exploits momentum and overlaps interaction stages to improve efficiency. When more time is available, it reallocates effort toward stabilizing contacts, reducing manipulation noise, and improving robustness under environmental mismatch. These behaviors arise not from hand-designed control rules but from learning how temporal constraints shape the trade-off between speed, stability, and task progress.

Our results suggest that explicit temporal reasoning can serve as a unifying mechanism linking several desirable properties of robot behavior. By making time observable and controllable, the policy learns to allocate temporal resources in a manner analogous to human action strategies. Humans routinely adjust tempo depending on urgency and precision requirements. The time-aware policy exhibits similar behavior, accelerating execution under tight schedules and slowing down to improve stability when additional time is available. In this sense, time functions as a behavioral regulator that links performance, robustness, and interaction strategy.

One practical implication of this formulation is improved robustness under sim-to-real discrepancies. When real-world dynamics differ from those seen during simulation training, policies often fail because their learned interaction strategies become unstable. Instead of requiring retraining or system identification, the time-aware policy can adapt by modulating its internal temporal progression. Allocating additional time allows the policy to re-stabilize contact interactions and maintain task success. This temporal adaptation acts as a meta-stabilization mechanism that enables the same learned policy to remain functional across environmental shifts.

Explicit temporal signals also facilitate coordination and human interaction. In multi-agent settings, predictable timing enables agents to synchronize actions and maintain punctual collaboration. Our experiments demonstrate that temporal reasoning allows policies to recover from disturbances while still meeting coordination schedules. Furthermore, the time-ratio interface provides a simple yet expressive mechanism for human users to modulate robot behavior online. Increasing the ratio induces faster and more dynamic execution, whereas decreasing it promotes cautious and stable interaction. Importantly, this control channel operates at the level of temporal regulation rather than low-level actuation, allowing users to influence robot behavior without modifying control policies or reward functions.

Our analysis further suggests that time awareness alters how the policy interprets sensory information. The learned policy dynamically changes which observations dominate its decision process depending on temporal context. Under urgent schedules, the policy focuses on variables directly related to task progress. Under relaxed schedules, it incorporates additional spatial and relational cues to improve precision and stability. These findings indicate that temporal signals can reorganize the internal representation of state and action selection, linking perception, control, and temporal reasoning within a unified policy.

Several limitations of the current work point toward directions for future research. First, our instability metric based on the sum of object linear velocities provides only a coarse proxy for contact stability. Richer measures incorporating force sensing, acceleration, or learned stability predictors may provide more accurate guidance during training. Second, each task in this study was trained with a separate time-aware policy. Extending the framework to multi-task or hierarchical learning could enable shared temporal representations that accelerate adaptation across tasks. Such representations may provide a promising pathway toward transferable temporal skills.

Third, our experiments rely primarily on state-based observations. Integrating time-aware policy learning with visual or multimodal perception may improve robustness in partially observable environments and enable deployment in more complex real-world settings. Finally, although the time-ratio interface offers an intuitive control channel, understanding how human operators interpret and use temporal modulation remains an open question. Systematic human studies could provide insights into how temporal control can best support collaborative human-robot interaction.

Beyond manipulation, the principle of time-aware control may apply broadly across robotic domains where timing and coordination are essential. Locomotion, aerial coordination, and assistive robotics all involve tasks in which agents must allocate effort across time while balancing speed and stability. Treating time as an explicit control dimension may therefore provide a general framework for enabling robots to regulate their behavior under varying temporal constraints.

More broadly, this work highlights a conceptual shift in robot learning. Rather than optimizing policies for a single fixed temporal regime, policies can learn to regulate how they use time itself. By learning not only \textit{what} action to execute but also \textit{when} and \textit{how fast} to execute it, robots can achieve behaviors that are efficient, robust, and naturally aligned with the temporal structure of real-world interaction. This perspective suggests that temporal reasoning may constitute a fundamental axis of robot intelligence, enabling systems that adapt their tempo, coordination, and precision in continuously changing environments.

\mysection{Materials and Methods}
\label{sec:methods}

\mysubsubsection{Method overview}

Time-aware policy learning aims to enable a robot policy to regulate its behavior according to temporal context while preserving task competence. The overall pipeline is illustrated in Fig.~\ref{fig:method_overview}. The framework proceeds in three stages.

First, we train a standard task-success policy $\pi_{\text{vanilla}}$ using a conventional reinforcement learning objective that maximizes expected discounted reward:
\begin{equation}
J(\pi)=\mathbb{E}\left[\sum_{t=0}^{\infty}\gamma^t r_t\right].
\label{eq:mdp_obj}
\end{equation}
This stage allows the agent to acquire the basic manipulation skills required to solve the task without introducing time pressure during early learning.

Second, we refine the policy to obtain a time-optimal policy $\pi^*_{\text{time}}$. The goal of this stage is not to produce the final deployable policy, but to estimate the fastest feasible behavior under the system dynamics. The resulting policy is used to compute two task-dependent reference quantities: the temporal lower bound $T^{\text{min}}(\phi)$ and the maximum manipulation instability $p^{\max}(\phi)$ for each environment configuration $\phi$.

Third, we train the time-aware policy. The time-optimal policy is first distilled into a temporally augmented policy through behavior cloning so that the policy can interpret temporal signals without losing previously acquired skills. The augmented policy is then optimized with a punctuality objective and a stability constraint within a constrained Markov decision process (CMDP). This stage enables the policy to adapt its behavior across temporal regimes while maintaining task success.

After training, the resulting policy receives both the system state and temporal signals $(T^{\text{left}}, tr)$ and produces actions conditioned on the temporal context. At inference time, the same policy can be deployed across different schedules by adjusting the time ratio $tr$ without retraining.

\mysubsubsection{Temporal representation}

The policy observes two temporal variables: the remaining time $T^{\text{left}}$ and the time ratio $tr$. The remaining time represents the remaining temporal budget for completing the task, while the time ratio modulates the internal progression of time perceived by the policy.

The remaining time evolves according to
\begin{equation}
T^{\text{left}}_{t+1}=T^{\text{left}}_t-\Delta t \cdot tr_t,
\end{equation}
where $\Delta t$ is the control interval. The time ratio therefore determines how quickly the internal clock progresses relative to real time. We illustrate the motivation for controlling the time elapse rather than the remaining time in Fig.~\ref{fig:scale_timepass}.

Decoupling $T^{\text{left}}$ and $tr$ improves learning stability and interpretability. Directly manipulating remaining time would introduce large discontinuities when schedules change. In contrast, the time ratio provides a normalized control signal that allows smooth modulation of behavior. In addition, explicitly including both variables in the observation preserves the Markov property of the decision process. Without this separation, the evolution of the temporal state would depend on hidden variables and require recurrent policies to infer temporal progress.

\mysubsubsection{Learning the temporal lower bound}
\label{sec:retrieve_timeoptimal}

The purpose of this stage is to estimate the fastest feasible successful behavior for a given task and environment configuration. A naive way to encourage fast completion is to rely on sparse terminal success reward with discounting. Let $\mathcal{S}^{\text{succ}}\subseteq\mathcal{S}$ denote the set of success states. The discrete success step is
\begin{equation}
t^{\text{succ}} \triangleq \inf\{t\in\mathbb{N}: s_t\in \mathcal{S}^{\text{succ}}\},
\end{equation}
and the corresponding completion time is $T^{\text{succ}} = t^{\text{succ}}\Delta t$.

A rollout-truncated success objective takes the form
\begin{equation}
J_R(\pi)
=
\mathbb{E}\!\left[
\mathbbm{1}_{\text{succ}}\,\gamma^{T^{\text{succ}}}R
+
\mathbbm{1}_{\text{trunc}}\,\gamma^{N}V_R^\pi(s_N)
\right],
\end{equation}
where $\mathbbm{1}_{\text{succ}}$ indicates successful completion before timeout, $\mathbbm{1}_{\text{trunc}}$ indicates truncation at step $N$, and $V_R^\pi(s_N)$ denotes the value of the truncated state. Although this objective prefers faster success in principle, in practice it is highly sensitive to the discount factor~\cite{sutton1998reinforcement} and often provides weak learning signals for long-horizon manipulation.

We therefore refine the objective by augmenting the terminal success reward with the remaining time at completion:
\begin{equation}
J_R(\pi)=
\mathbb{E}\!\left[
\mathbbm{1}_{\text{succ}}\cdot \gamma^{T^{\text{succ}}}(R\cdot T^{\text{left}}_K)
+
\mathbbm{1}_{\text{trunc}}\cdot \gamma^{N}V_R^\pi(s_N)
\right],
\quad
T^{\text{left}}_K=\max(N\Delta t-T^{\text{succ}},0),
\label{eq:timeoptimal_obj}
\end{equation}
where $T^{\text{left}}_K$ denotes the remaining time at the terminal step $K$. This objective explicitly favors successful trajectories that finish with greater remaining time and therefore yields a stronger signal for temporal efficiency. Because learning such an objective from scratch is difficult in sparse-reward, long-horizon tasks, we initialize this stage from the previously trained vanilla policy and continue optimization from there. The resulting policy is denoted $\pi^*_{\text{time}}$.

Executing $\pi^*_{\text{time}}$ over randomized environment configurations yields the empirical temporal lower bound
\begin{equation}
T^{\text{min}}(\phi)=
\mathbb{E}\!\left[T^{\text{succ}} \mid \pi^*_{\text{time}},\phi\right],
\end{equation}
which serves as the feasible minimum completion time for the corresponding task configuration $\phi$.

In addition to the temporal lower bound, we estimate the maximum manipulation instability observed during efficient execution. Scene instability is defined as
\begin{equation}
p_t=\sum_i \left\|v_t^i\right\|_2,
\label{eq:instability}
\end{equation}
where $v_t^i$ denotes the translational velocity of the $i$-th movable object. The peak instability during a successful rollout is
\begin{equation}
p^{\text{peak}}=\max_{0\le t\le T^{\text{succ}}} p_t,
\end{equation}
and the corresponding reference instability is estimated as
\begin{equation}
p^{\max}(\phi)=
\mathbb{E}\!\left[p^{\text{peak}} \mid \pi^*_{\text{time}},\phi\right].
\end{equation}

$(T^{\text{min}}(\phi), p^{\max}(\phi))$ provide physically grounded references for the subsequent time-aware training stage.

\mysubsubsection{Embedding temporal observations}

Before optimizing the final time-aware objective, we introduce temporal signals into the policy using behavior cloning. This step ensures that the temporally augmented policy can interpret the additional temporal inputs while preserving the efficient manipulation behavior learned by $\pi^*_{\text{time}}$.

The cloning objective is
\begin{equation}
L^{\text{BC}}(\pi)=
\mathbb{E}_{a\sim \pi^*_{\text{time}}(\cdot|s)}
\left[
\log \pi^*_{\text{time}}(a|s)-
\log \pi(a|s,T^{\text{left}},tr)
\right].
\label{eq:TeacherStudent}
\end{equation}

During this stage, the time ratio is sampled uniformly from $[0.2,1]$ at the beginning of each episode and remains fixed within the episode. The remaining time is initialized from the task-specific temporal reference and evolves according to Eq.~\ref{eq:TimeAwareObj}. Training continues until the augmented policy reproduces the performance of $\pi^*_{\text{time}}$ without substantial degradation.

\mysubsubsection{Time-aware policy optimization}
\label{sec:retrieve_timeaware}

The final time-aware policy is trained using a constrained Markov decision process (MDP) that balances task success, punctuality, and stability.

The policy takes the form
\begin{equation}
\pi(a_t \mid s_t,T_t^{\text{left}},tr_t),
\end{equation}
and is optimized with a sparse success reward penalized by timing mismatch:
\begin{equation}
J_R(\pi)=
\mathbb{E}\!\left[
\mathbbm{1}_{\text{succ}}\cdot \max(R-R_t\delta^{\text{time}},0)
+
\mathbbm{1}_{\text{trunc}}\cdot V_R^\pi(s_N)
\right],
\end{equation}
where
\begin{equation}
\delta_{\text{time}}=
\left|\frac{T^{\text{left}}_K}{tr_K}\right|.
\end{equation}

The division by $tr_K$ is necessary because $T^{\text{left}}$ evolves in the policy's internal clock rather than real time. Dividing by $tr_K$ converts the terminal residual back into the corresponding real-time mismatch, ensuring that the punctuality penalty is comparable across different temporal regimes.

To ensure effective use of available time, we impose an instability constraint
\begin{equation}
c_t^{\text{inst}}=
\max\bigl(p_t-p^{\max}\cdot tr_t,0\bigr),
\label{eq:InstabilityCost}
\end{equation}
which allows more dynamic manipulation under urgent schedules and encourages gentler behavior when more time is available.

The resulting constrained objective is
\begin{equation}
\max_\pi J_R(\pi)
\quad
\text{s.t.}
\quad
J_C(\pi)\le \epsilon,
\end{equation}
where
\begin{equation}
J_C(\pi)=
\mathbb{E}\left[
\sum_{t=0}^{\infty}\gamma_c^t c_t^{\text{inst}}
+
\mathbbm{1}_{\text{trunc}}\cdot \gamma_c^N V_C^\pi(s_N)
\right].
\label{eq:TimeawareObjFunc}
\end{equation}

This formulation avoids manually tuning a weighted trade-off between punctuality and stability, and allows separate reward and cost critics to estimate value and constraint signals accurately.

Following prior works \cite{altman2021constrained, zhang2022penalized}, this constrained policy optimization problem can be reformulated as:
\begin{equation}
\begin{aligned}
    & \max_{\pi'} \quad \mathbb{E} \left[ A_{R,t}^{\pi}(s, a) \right] \\
    & \text{s.t.} \quad \underbrace{J_{C_i} (\pi) + \frac{1}{1-\gamma} \mathbb{E}_{\pi'} \left[ A_{C_i,t}^{\pi}(s, a) \right]}_{J_{C_i}(\pi')} \leq \epsilon_i, \; \forall i,
\end{aligned}
\end{equation}
where $\pi$ denotes the current policy,  $\pi'$ denotes the updated policy, and $A_{R,t}^{\pi}(s,a)$ and $A_{C_i,t}^{\pi}(s,a)$ are unbiased estimators of the reward advantage and the $i$-th cost advantage at time step $t$, respectively.

\mysubsubsection{Policy optimization}
\label{sec:cmdp_formulation}

We optimize the constrained objective using Penalized Proximal Policy Optimization (P3O)~\cite{zhang2022penalized}, which extends PPO~\cite{schulman2017proximal} with explicit penalties for constraint violations. The P3O objective is
\begin{equation}
L^{\text{P3O}}(\theta')
=
L_R^{\text{CLIP}}(\theta')
-
\sum_i \kappa_i \max\{0,L_{C_i}^{\text{VIOL}}(\theta')\},
\end{equation}
where $\kappa_i$ controls the penalty strength for each constraint.

The reward clipping objective is
\begin{equation}
L_R^{\text{CLIP}}(\theta')
=
\mathbb{E}_t
\left[
\min\bigl(r_t(\theta')A_{R,t}^{\pi_\theta},
\mathbf{Clip}(r_t(\theta'))A_{R,t}^{\pi_\theta}\bigr)
\right],
\end{equation}
where
\[
r_t(\theta')=\frac{\pi_{\theta'}(a_t|s_t)}{\pi_\theta(a_t|s_t)}.
\]

The cost violation term is
\begin{align}
L_{C_i}^{\text{VIOL}}(\theta')
&=
L_{C_i}^{\text{CLIP}}(\theta')
+
(1-\gamma_c)\bigl(J_{C_i}(\pi_\theta)-\epsilon_i\bigr),\\
L_{C_i}^{\text{CLIP}}(\theta')
&=
\mathbb{E}_t
\left[
\max\bigl(r_t(\theta')A_{C_i,t}^{\pi_\theta},
\mathbf{Clip}(r_t(\theta'))A_{C_i,t}^{\pi_\theta}\bigr)
\right].
\end{align}

Since the unnormalized $A^{\pi_\theta}_{R,t}$ and $A^{\pi_\theta}_{C_i,t}$ have different magnitudes, direct optimization may bias updates toward objectives with larger numerical scales. We therefore normalize both advantages to improve the policy learning. To ensure the constraint objective (\ie $\mathbb{E}_t \left[A^{\pi_\theta}_{C_i,t} \right] + (1 - \gamma_c)\big( J_{C_i}(\pi_\theta)-\epsilon_i \big) \leq 0$) remains unchanged, we adopt the same cost objective from prior work \cite{lee2023evaluation}:

\begin{align}
L^{\text{N-VIOL}}_{C_i}(\theta')
&= L^{\text{N-CLIP}}_{C_i}(\theta') 
+ \frac{(1 - \gamma_c)\big(J_{C_i}(\pi_\theta) - \epsilon_i \big)+\mu_{C_i}}{\sigma_{C_i}}, \\
L^{\text{N-CLIP}}_{C_i}(\theta') 
&= \mathbb{E}_t \left[
     \max \Big( r_t(\theta') \tilde{A}^{\pi_\theta}_{C_i,t},\;
                \mathbf{Clip}(r_t(\theta')) \tilde{A}^{\pi_\theta}_{C_i,t}
        \Big)
    \right], \\
& \quad\quad \tilde{A}^{\pi_\theta}_{C_i,t} 
= \frac{A^{\pi_\theta}_{C_i,t}-\mu_{C_i}}{\sigma_{C_i}},
\end{align}
where $\mu_{C_i}$ and $\sigma_{C_i}$ are the mean and standard deviations of $A^{\pi_\theta}_{C_i,t}$. This normalization keeps the optimization scale consistent and allows the policy to learn stably across multiple cost constraints.

Therefore, the normalized P3O objective then becomes:
\begin{equation}
L^{\text{N-P3O}}(\theta') 
= L^{\text{N-CLIP}}_{R}(\theta') 
  - \sum_{i} \kappa_i \cdot \max \left\{ 0,\; L^{\text{N-VIOL}}_{C_i}(\theta') \right\},
\end{equation}

The final loss combines the normalized P3O objective with value and cost critic regressions and an entropy bonus to encourage policy exploration:
\begin{equation}
L(\theta') = c_1 L^{\text{C-V}}(\theta') + c_2 L^{\text{C-C}}(\theta') - c_3 L^{\text{H}}(\theta') -  L^{\text{N-P3O}}(\theta')
\end{equation}
with
\begin{gather}
    L^{\text{C-V}}(\theta') = \mathbb{E}_t \left[ \left( V^R_{\theta'}(s_t) - G^R_t \right)^2 \right], \\
    L^{\text{C-C}}(\theta') = \mathbb{E}_t \left[ \left( V^C_{\theta'}(s_t) - G^C_t \right)^2 \right], \\
    L^{\text{H}}(\theta') = \mathbb{E}_t \left[ \mathcal{H}[\pi_{\theta'}](s_t) \right],
\end{gather}
where $V^R_{\theta'}(s_t)$ is the critic for the state value, $V^C_{\theta'}(s_t)$ is the critic for the state cost, $G^R_t$ is the cumulative return, $G^C_t$ is the cumulative cost, and $\mathcal{H}[\pi_{\theta'}](s_t)$ is the policy entropy at state $s_t$. The separate critics ensure accurate value and constraint estimation, and the entropy term maintains action diversity to prevent premature convergence.

\mysubsubsection{Policy architecture}

The actor, reward critic, and cost critic are implemented as five-layer multilayer perceptrons with identical hidden dimensions but independent parameters. The actor receives task observations, robot state, temporal variables $(T_t^{\mathrm{left}}, tr_t)$, and previous actions. It outputs the parameters of a Beta distribution for six continuous end-effector motion commands and one binary gripper command. To improve sim-to-real transfer, we use an asymmetric actor--critic design. The critics receive privileged information, including precise instability values and task-specific simulator states, while the actor operates only on noisy observations available at deployment.

\mysubsubsection{Training procedure}

Training proceeds in three stages corresponding to the vanilla policy, the time-optimal policy, and the final time-aware policy. For time-optimal training, the actor from the vanilla policy is reused while the critic is reinitialized so that value estimation remains accurate under the modified objective~\cite{xiao2025efficient, choars, sun2022exploit, ahn2024reset, wolczyk2024fine}. During the initial iterations, only the critic is updated to stabilize value prediction before joint actor--critic updates resume. PPO is used with a stricter KL-divergence threshold to prevent policy collapse in the sparse-reward setting. For temporal embedding, the time-optimal policy is distilled into an augmented student policy through behavior cloning with sampled time ratios. For the final time-aware stage, environment configurations are sampled together with their corresponding $(T^{\text{min}},p^{\max})$ estimates. The time ratio is sampled uniformly from $[0.2,1]$ and held constant within the episode, while remaining time evolves according to Eq.~\ref{eq:TimeAwareObj}. P3O is then used to optimize the final policy.

\mysubsubsection{Sim-to-real transfer}

To facilitate sim-to-real transfer, joint-level system identification is performed before training. Joint limits, velocity bounds, torque limits, and actuator delays are measured on the physical robot and incorporated into simulation. Observation noise is injected during training to improve robustness. Detailed noise settings are provided in the Supplementary Material Table~\ref{tab:noise_level}. The policy runs at 20\,Hz for the arm and 2\,Hz for the gripper, while the low-level joint impedance controller runs at 1\,kHz. During deployment, commanded joint positions are computed incrementally from previous commands to mitigate latency-induced drift.

\mysubsubsection{Heuristic stage-wise control details}
\label{sec:stage_wise_control}

For heuristic stage-wise control, a task is divided into semantic stages and each stage is assigned either an efficient (E) or stable (S) mode with time proportion $P_i$. To ensure the total scheduled duration remained constant and all time ratios stay within valid bounds, we first compute the ratio of allocated time proportions between stable and efficient stages:
\begin{equation}
k=
\frac{\sum_{i=1}^{M}\mathbbm{1}[\text{Stage}_i=\text{S}]P_i}
{\sum_{i=1}^{M}\mathbbm{1}[\text{Stage}_i=\text{E}]P_i},
\end{equation}
where $M$ is the number of stages.

Given
\begin{equation}
\bar{tr}=\frac{T^{\text{min}}}{T^{\text{goal}}},
\end{equation}
the stable and efficient time ratios are
\begin{equation}
\Delta tr_S=
\min\left(
\bar{tr}-tr_{\min},
\frac{tr_{\max}-\bar{tr}}{k}
\right),
\qquad
\Delta tr_E=k\Delta tr_S,
\end{equation}
and
\begin{equation}
tr_E=\bar{tr}+\Delta tr_E,
\qquad
tr_S=\bar{tr}-\Delta tr_S.
\end{equation}

This construction ensures that stable stages receive smaller time ratios and efficient stages larger ones while preserving the overall schedule. The complete stage definitions for all tasks are listed in Table~\ref{tab:Staged_wise_control}.


\clearpage 

%
\bibliography{science_template} 
\bibliographystyle{sciencemag}

%
%
%
%
%
%


\section*{Acknowledgments}

\paragraph*{Funding:} 
This work is supported by DARPA TIAMAT program under award HR00112490419, ARO under award W911NF2410405, and ARL STRONG program under awards W911NF2320182, W911NF2220113, and W911NF242021.

\paragraph*{Author contributions:}
B.C., Y.J. conceived and designed the research. Y.J. performed simulations and physical experiments. All authors analyzed data and wrote the manuscript.

\paragraph*{Competing interests:}
Duke University has filed patent rights for the technology associated with this manuscript. For further license rights, including using the patent rights for commercial purposes, please contact Duke's Office for Translation and Commercialization (otcquestions@duke.edu) and reference OTC DU9041PROV.

\paragraph*{Data and materials availability:}
The code is available at \url{https://drive.google.com/drive/folders/1sCfGEupQoO3WC7Iq4Qb0ysiXgS5o5L7w?usp=drive_link}.

The high-resolution version of all the supplementary videos can be accessed at \url{https://drive.google.com/drive/folders/1UZq4122AkPZ0vdQmO9ffB7DzUEUgZ97c?usp=drive_link}.


\mysubsection{Supplementary materials}
Materials and Methods\\
Supplementary Text\\
Tables S1 to S4\\
Figs. S1 to S4 \\
\textbf{Supplementary Movie S1 --- Overview}: Overview of the motivation, key idea, and demonstrations.\\
\textbf{Supplementary Movie S2 --- Baseline Interpolation}: Direct interpolation of vanilla policy without temporal awareness.\\
\textbf{Supplementary Movie S3 --- Optimized Efficiency}: Time awareness enables faster task completion while maintaining punctuality at minimum scheduled time (${tr}_t = 1.0$).\\
\textbf{Supplementary Movie S4 --- Flexible Punctuality}: Time awareness adapts behavior to meet variable deadlines across different scheduled times ($0 < {tr}_t < 1$).\\
\textbf{Supplementary Movie S5 --- Robust Adaptation}: Time awareness enhances stability and performance under varying environmental conditions.\\
\textbf{Supplementary Movie S6 --- Coordinated Delivery}: Multiple agents collaboratively transport objects using time-aware coordination.\\
\textbf{Supplementary Movie S7 --- Interactive Control}: Human-in-the-loop temporal control for real-time behavior adaptation.\\


\newpage


\renewcommand{\thefigure}{S\arabic{figure}}
\renewcommand{\thetable}{S\arabic{table}}
\renewcommand{\theequation}{S\arabic{equation}}
\renewcommand{\thepage}{S\arabic{page}}
\setcounter{figure}{0}
\setcounter{table}{0}
\setcounter{equation}{0}
\setcounter{page}{1} 


\begin{center}
\mysection{Supplementary Materials for\\ \scititle}


Yinsen Jia$^{1}$,
Boyuan Chen$^{1, 2, 3\ast}$\\
\small$^{1}$Department of Electrical and Computer Engineering, Duke University\\
    \small$^{2}$Department of Mechanical Engineering and Materials Science, Duke University\\
    \small$^{3}$Department of Computer Science, Duke University\\
\small$^\ast$To whom correspondence should be addressed; Email: boyuan.chen@duke.edu.
\end{center}

\subsubsection*{This PDF file includes:}
Materials and Methods\\
Supplementary Text\\
Tables S1 to S4\\
Figures S1 to S4\\

\newpage
\mysubsection{Materials and Methods}
\label{appendix:methods}

\begin{table}[h!]
	\centering
	\caption{\textbf{Stage-Wise Control Parameters for Different Tasks.}
            This table lists, for each task, the sequence of stages, the stage modes, the relative time portion per stage, and the total scheduled time. Efficient mode (E) encourages the policy to act faster, and stable mode (S) encourages the policy to act more cautiously. The time portion values represent the proportion of the total scheduled time allocated to each stage.}
	\label{tab:Staged_wise_control}
	\footnotesize  
	\begin{tabular}{lcccc}
		\\
		\hline
		Task Name & Stage Name & Stage Mode & Time Portion & Scheduled Time\\
		\hline
		Cube Stacking & Approach, Grasp, Transport, Stack & E, S, E, S & 0.15, 0.35, 0.15, 0.35 & $2T^{\text{min}}$\\
		Granular Pouring & Transport, Pour & E, S & 0.5, 0.5 & $2T^{\text{min}}$\\
		Drawer Opening & Approach, Grasp, Pull, Open & E, S, E, S & 0.2, 0.2, 0.3, 0.3 & $2T^{\text{min}}$\\
		\hline
	\end{tabular}
\end{table}

\begin{table}
	\centering
	\caption{\textbf{Default Training Hyperparameters.}
		This table lists the hyperparameters used for training the policy with PPO algorithm, including their values and descriptions.}
	\label{tab:ppo_hyper}
	\small
	\begin{tabular}{lcc}
		\\
		\hline
		Hyperparameter & Value & Description\\
		\hline
		Learning Rate & 0.0002 & Learning rate\\
		Num Envs & 16384 & The number of parallel simulation envs\\
            $\Delta t$ & 1/20 & The control interval\\
		Roll-out Steps & 32 & The number of roll-out steps per update\\
		Batch Size & 131072 & Batch size\\
		Update Epochs & 5 & Update epochs\\
		$\gamma$ & 0.995 & The discount factor of the reward\\
		$\lambda_{gae}$ & 0.95 & The discount for GAE\\
		$\epsilon$ & 0.2 & The surrogate clipping coefficient\\
		$c_1$ & 0.5 & The coefficient of the value function\\
		$c_3$ & 0.005 & The coefficient of the entropy\\
		$g_{clip\_max}$ & 0.5 & Gradient clipping\\
            Target KL & 2.5 & Maximum KL before reverting the update\\
		Hidden Layer Size & [256, 128, 64] & The number of units\\
		$R$ & 1000 & Task Success reward\\
		\hline
	\end{tabular}
\end{table}

\clearpage
\begin{table}
	\centering
	\caption{\textbf{Time-Aware Training Stage Overrides.}
		This table lists the hyperparameter overrides specifically used for time-aware training stages, including their values and descriptions.}
	\label{tab:timeaware_hyper}
	\begin{tabular}{lcc}
		\\
		\hline
		Hyperparameter & Value & Description\\
		\hline
		$\gamma$ & 1.0 & The discount factor of the reward\\
		$\gamma_c$ & 0.99 & The discount factor of the cost\\
		$c_2$ & 0.5 & The coefficient of the cost function\\
		$R_t$ & 100 & The reward scale of the punctuality objective\\
		\hline
	\end{tabular}
\end{table}

\clearpage
\begin{table}
	\centering
	\caption{\textbf{Details of Noise.}
		This table lists the noise levels applied to different observations across various task categories and robot states. The noise was uniformly sampled within the specified ranges.}
	\label{tab:noise_level}
	\small
	\begin{tabular}{llcc}
		\\
		\hline
		Category Name 
            & Observation Name & Unit & Level\\
		\hline
		Cube Stacking 
            & CubeA Pos & m & $\mathcal{U}(\pm 0.01)$\\
		  & CubeA Rot & rad & $\mathcal{U}(\pm \pi/60)$\\
		  & CubeB Pos & m & $\mathcal{U}(\pm 0.01)$\\
		  & CubeB Rot & rad & $\mathcal{U}(\pm \pi/60)$\\
		\hline
		GM Pouring 
            & CupA Pos & m & $\mathcal{U}(\pm 0.01)$\\
		  & CupA Rot & rad & $\mathcal{U}(\pm \pi/60)$\\
		  & CupB Pos & m & $\mathcal{U}(\pm 0.01)$\\
		  & CupB Rot & rad & $\mathcal{U}(\pm \pi/60)$\\
		\hline
		Drawer Opening 
            & Handle Pos & m & $\mathcal{U}(\pm 0.01)$\\
		\hline
		Robot States 
            & EEF Pos & m & $\mathcal{U}(\pm 0.01)$\\
		  & EEF Rot & rad & $\mathcal{U}(\pm \pi/60)$\\
		  & Joint Pos & rad & $\mathcal{U}(\pm \pi/60)$\\
            \hline
            Robot Controller
		  & Gripper Velocity & m/s & $\mathcal{U}(\pm 0.005)$\\
		  & Gripper Delay & s & $\mathcal{U}(0.1, 0.3)$\\
		\hline
	\end{tabular}
\end{table}

\begin{figure}
    \centering
    \includegraphics[width=\linewidth]{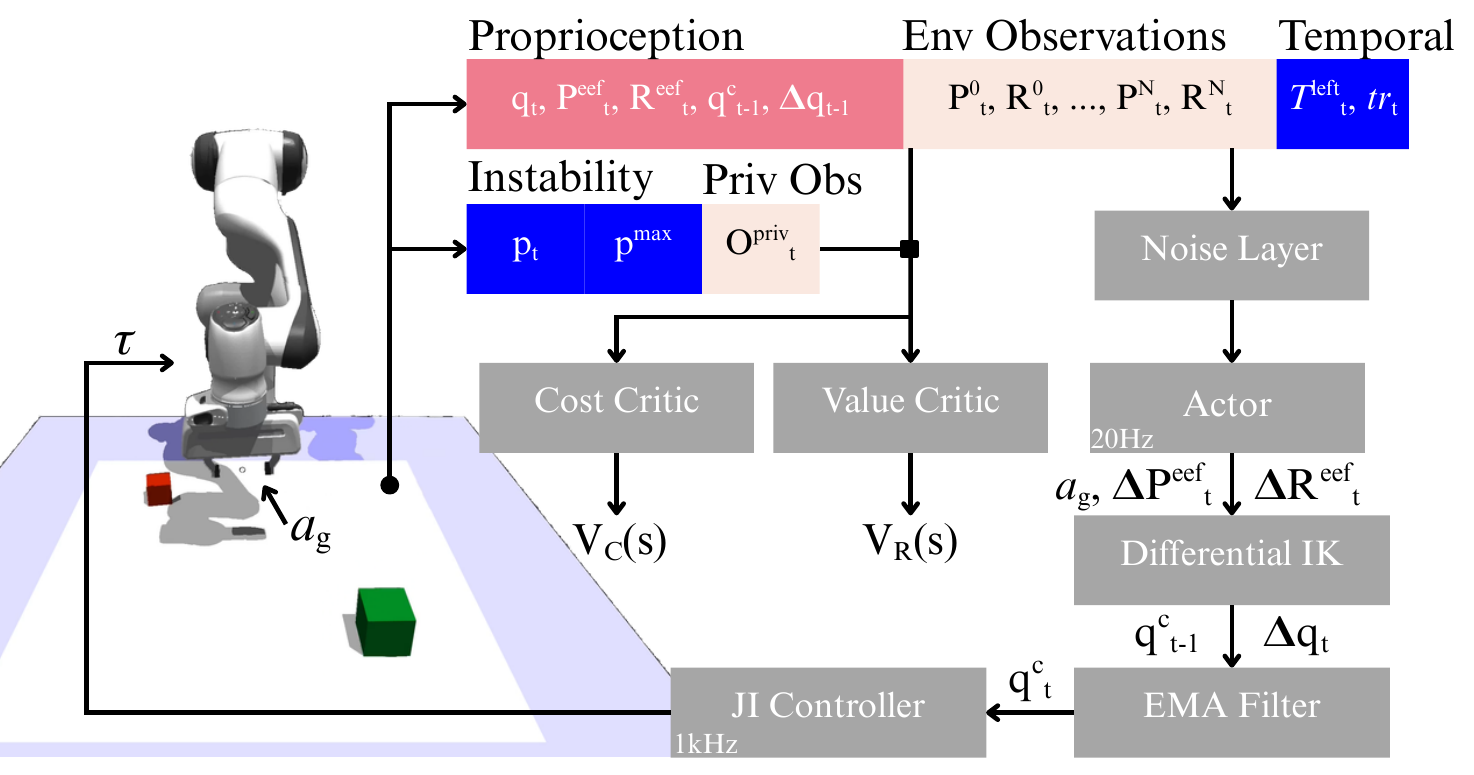}
    \caption{
        \textbf{Policy architecture and control loop.} During training, observations are passed through a noise layer before being input to the actor, which outputs 6-DoF end-effector relative poses and binary gripper actions at 20 Hz. These are converted to target joint positions via differential inverse kinematics, smoothed with an exponential moving average filter, and sent to the low-level joint impedance controller operating at 1 kHz. To address the challenge of measuring scene instability in real-world deployments, we employ an asymmetric actor–critic method. The critic observes noise-free observations and privileged information, while the actor uses only noisy and observable inputs. This design enables the actor to learn a deployable policy under partial observability while the critic leverages privileged information for accurate gradient shaping.
    }
    \label{fig:model_architecture}
\end{figure}

\begin{figure}
    \centering
    \includegraphics[width=\linewidth]{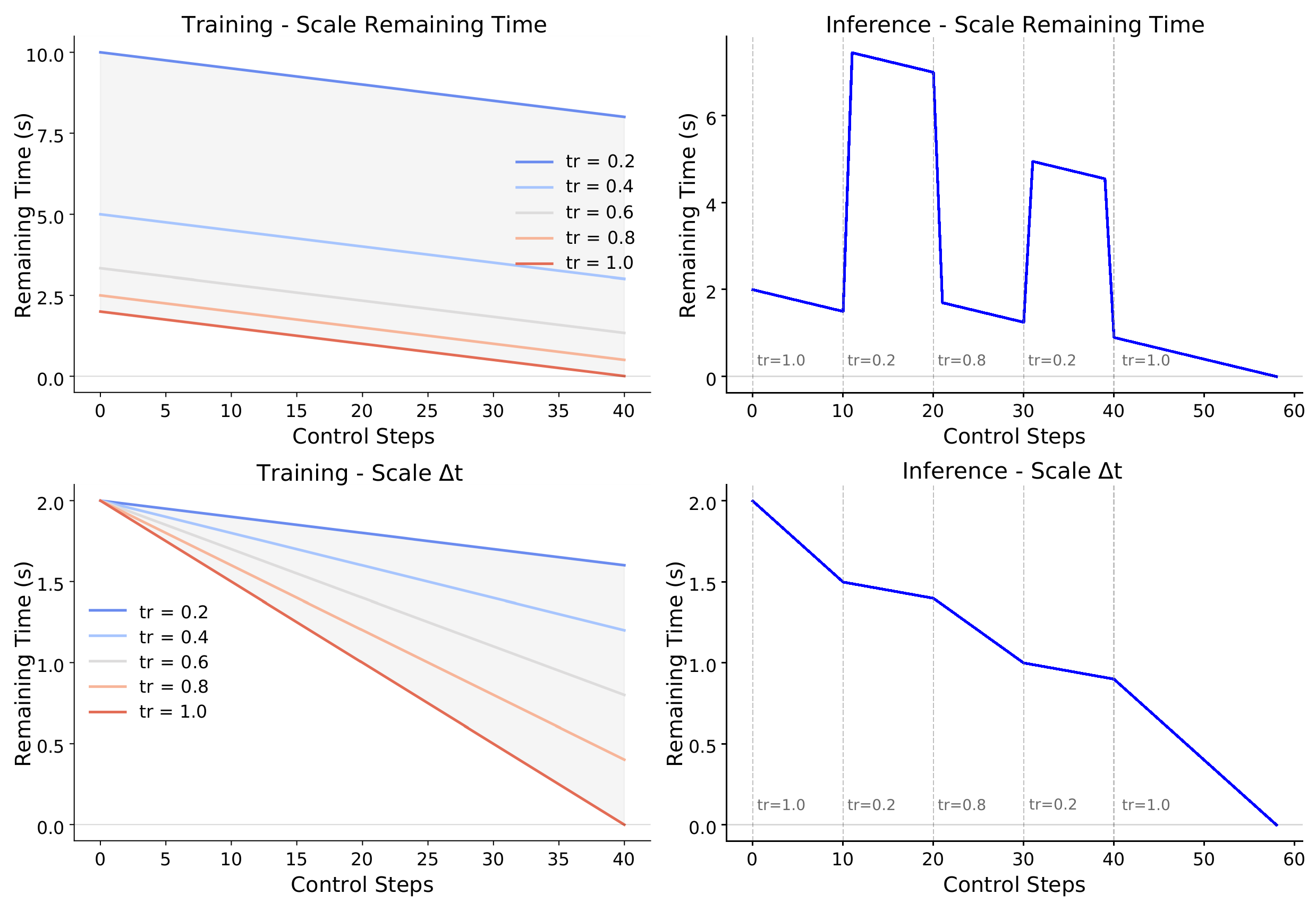}
    \caption{
        \textbf{Controlling time elapse via the time ratio simplifies training and stabilizes inference.}
        We compare two ways of applying the time ratio: directly scaling the remaining time versus scaling the time elapse, using a toy example. In this example, we set the temporal lower bound to 2\,s, $\Delta t=0.05$\,s, and use a pre-defined time-ratio control sequence. Directly scaling the remaining time yields a fixed time elapse during training. However, if the same scaling is applied at inference to modulate the policy behavior, changes in the time ratio can induce large, previously unseen oscillations in the remaining time. These oscillations introduce a strong distribution shift between training and inference, which can easily cause the policy to collapse. In contrast, scaling the time elapse produces different slopes for the remaining-time trajectory. At inference, this results in a smooth, monotonically decreasing remaining time observation. Consequently, even without step-by-step randomization of the time ratio during training, the policy can easily interpolate across slope changes at inference and adapt its behaviors.
    }
    \label{fig:scale_timepass}
\end{figure}

\begin{figure}
    \centering
    \includegraphics[width=\linewidth]{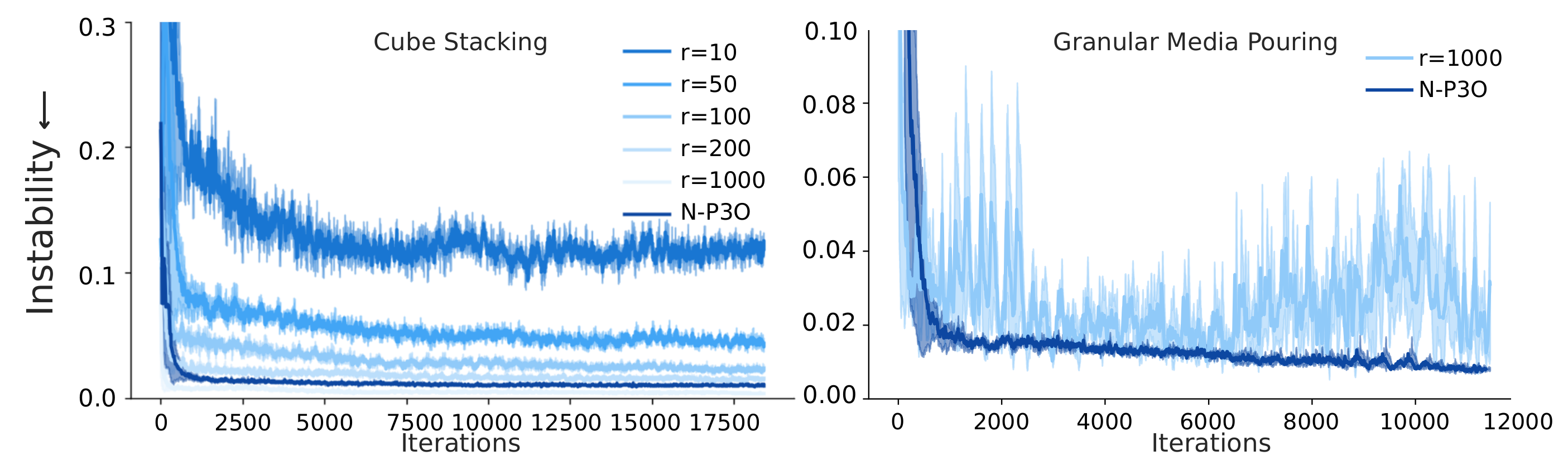}
    \caption{
        \textbf{CMDP stabilizes time-aware training across tasks.} 
        We evaluated the effect of removing our constrained MDP (Method:~\hyperref[sec:cmdp_formulation]{Policy optimization}) formulation by directly applying a dense instability penalty on reward. Since the resulting objective combined sparse task rewards, time rewards, and an instability penalty term, we had to manually tune penalty scalings in this study. We searched five different scalings of the penalty, and each reward combination was trained with 3 different seeds. Without CMDP formulation, time-aware policy training becomes highly sensitive to reward scaling. Large penalty scaling reduced object interaction and could even break policy training, while small scaling failed to enforce the instability constraint effectively. Moreover, this tuning was required for each task. For example, cube stacking requires sufficiently high instability reward scaling, but this same scaling destabilizes granular media pouring training entirely. In contrast, our CMDP formulation automatically maintained stability without per-task tuning by treating instability as a constraint during policy optimization.
    }
    \label{fig:cmdp_ablation}
\end{figure}

\begin{figure}
    \centering
    \includegraphics[width=\linewidth]{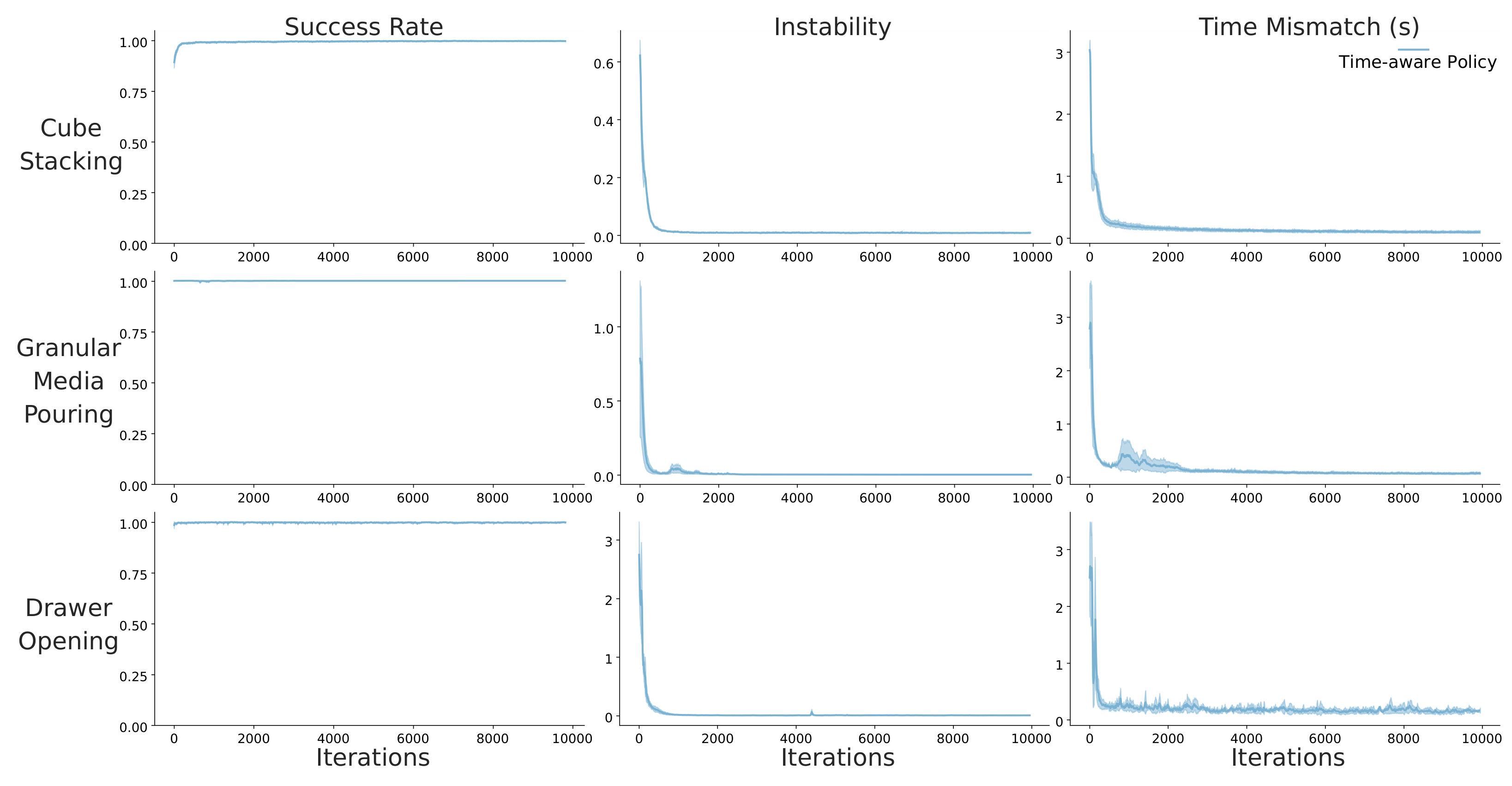}
    \caption{
        \textbf{Training curves of the time-aware policy.} Each experiment was conducted with five random seeds. The blue curve represents the mean across five training runs; the shaded region represents the standard deviation. The time-aware policy learning framework demonstrates strong training stability and convergence.
    }
    \label{fig:training_curve}
    \vspace{-2mm}
\end{figure}

\end{document}